# From random walks to distances on unweighted graphs

**Tatsunori B. Hashimoto**
MIT EECS
thashim@mit.edu

**Yi Sun**
MIT Mathematics
yisun@mit.edu

**Tommi S. Jaakkola**
MIT EECS
tommi@mit.edu

## Abstract

Large unweighted directed graphs are commonly used to capture relations between entities. A fundamental problem in the analysis of such networks is to properly define the similarity or dissimilarity between any two vertices. Despite the significance of this problem, statistical characterization of the proposed metrics has been limited.

We introduce and develop a class of techniques for analyzing random walks on graphs using stochastic calculus. Using these techniques we generalize results on the degeneracy of hitting times and analyze a metric based on the Laplace transformed hitting time (LTHT). The metric serves as a natural, provably well-behaved alternative to the expected hitting time. We establish a general correspondence between hitting times of the Brownian motion and analogous hitting times on the graph. We show that the LTHT is consistent with respect to the underlying metric of a geometric graph, preserves clustering tendency, and remains robust against random addition of non-geometric edges. Tests on simulated and real-world data show that the LTHT matches theoretical predictions and outperforms alternatives.

## 1 Introduction

Many network metrics have been introduced to measure the similarity between any two vertices. Such metrics can be used for a variety of purposes, including uncovering missing edges or pruning spurious ones. Since the metrics tacitly assume that vertices lie in a latent (metric) space, one could expect that they also recover the underlying metric in some well-defined limit. Surprisingly, there are nearly no known results on this type of consistency. Indeed, it was recently shown [19] that the expected hitting time degenerates and does not measure any notion of distance.

We analyze an improved hitting-time metric – Laplace transformed hitting time (LTHT) – and rigorously evaluate its consistency, cluster-preservation, and robustness under a general network model which encapsulates the latent space assumption. This network model, specified in Section 2, posits that vertices lie in a latent metric space, and edges are drawn between nearby vertices in that space. To analyze the LTHT, we develop two key technical tools. We establish a correspondence between functionals of hitting time for random walks on graphs, on the one hand, and limiting Itô processes (Corollary 4.4) on the other. Moreover, we construct a weighted random walk on the graph whose limit is a Brownian motion (Corollary 4.1). We apply these tools to obtain three main results.

First, our Theorem 3.5 recapitulates and generalizes the result of [19] pertaining to degeneration of expected hitting time in the limit. Our proof is direct and demonstrates the broader applicability of the techniques to general random walk based algorithms. Second, we analyze the Laplace transformed hitting time as a one-parameter family of improved distance estimators based on random walks on the graph. We prove that there exists a scaling limit for the parameter $\beta$ such that the LTHT can become the shortest path distance (Theorem S5.2) or a consistent metric estimator averaging over many paths (Theorem 4.5). Finally, we prove that the LTHT captures the advantages of random-walk based metrics by respecting the cluster structure (Theorem 4.6) and robustly recovering similarity queries when the majority of edges carry no geometric information (Theorem 4.9). We now discuss the relation of our work to prior work on similarity estimation.

**Quasi-walk metrics:** There is a growing literature on graph metrics that attempts to correct the degeneracy of expected hitting time [19] by interpolating between expected hitting time and shortest path distance. The work closest to ours is the analysis of the phase transition of the $p$-resistance metric in [1] which proves that $p$-resistances are nondegenerate for some $p$; however, their work did not address consistency or bias of $p$-resistances. Other approaches to quasi-walk metrics such as logarithmic-forest [3], distributed routing distances [16], truncated hitting times [12], and randomized shortest paths [8, 21] exist but their statistical properties are unknown. Our paper is the first to prove consistency properties of a quasi-walk metric.

**Nonparametric statistics:** In the nonparametric statistics literature, the behavior of $k$-nearest neighbor and $\varepsilon$-ball graphs has been the focus of extensive study. For undirected graphs, Laplacian-based techniques have yielded consistency for clusters [18] and shortest paths [2] as well as the degeneracy of expected hitting time [19]. Algorithms for exactly embedding $k$-nearest neighbor graphs are similar and generate metric estimates, but require knowledge of the graph construction method, and their consistency properties are unknown [13]. Stochastic differential equation techniques similar to ours were applied to prove Laplacian convergence results in [17], while the process-level convergence was exploited in [6]. Our work advances the techniques of [6] by extracting more robust estimators from process-level information.

**Network analysis:** The task of predicting missing links in a graph, known as link prediction, is one of the most popular uses of similarity estimation. The survey [9] compares several common link prediction methods on synthetic benchmarks. The consistency of some local similarity metrics such as the number of shared neighbors was analyzed under a single generative model for graphs in [11]. Our results extend this analysis to a global, walk-based metric under weaker model assumptions.

## 2 Continuum limits of random walks on networks

### 2.1 Definition of a spatial graph

We take a generative approach to defining similarity between vertices. We suppose that each vertex $i$ of a graph is associated with a latent coordinate $x_i \in \mathbb{R}^d$ and that the probability of finding an edge between two vertices depends solely on their latent coordinates. In this model, given only the unweighted edge connectivity of a graph, we define natural distances between vertices as the distances between the latent coordinates $x_i$. Formally, let $\mathcal{X} = \{x_1, x_2, \ldots\} \subset \mathbb{R}^d$ be an infinite sequence of points drawn i.i.d. from a differentiable density with bounded log gradient $p(x)$ with compact support $D$. A spatial graph is defined by the following:

**Definition 2.1** (Spatial graph). *Let $\varepsilon_n : \mathcal{X}_n \to \mathbb{R}_{>0}$ be a local scale function and $h : \mathbb{R}_{\geq 0} \to [0, 1]$ a piecewise continuous function with $h(x) = 0$ for $x > 1$, $h(1) > 0$, and $h$ left-continuous at 1. The spatial graph $G_n$ corresponding to $\varepsilon_n$ and $h$ is the random graph with vertex set $\mathcal{X}_n$ and a directed edge from $x_i$ to $x_j$ with probability $p_{ij} = h(|x_i - x_j|\varepsilon_n(x_i)^{-1})$.*

This graph was proposed in [6] as the generalization of $k$-nearest neighbors to isotropic kernels. To make inference tractable, we focus on the large-graph, small-neighborhood limit as $n \to \infty$ and $\varepsilon_n(x) \to 0$. In particular, we will suppose that there exist scaling constants $g_n$ and a deterministic continuous function $\overline{\varepsilon} : D \to \mathbb{R}_{>0}$ so that

$$g_n \to 0, \qquad g_n n^{\frac{1}{d+2}} \log(n)^{-\frac{1}{d+2}} \to \infty, \qquad \varepsilon_n(x) g_n^{-1} \to \overline{\varepsilon}(x) \text{ for } x \in \mathcal{X}_n,$$

where the final convergence is uniform in $x$ and a.s. in the draw of $\mathcal{X}$. The scaling constant $g_n$ represents a bound on the asymptotic sparsity of the graph.

We give a few concrete examples to make the quantities $h$, $g_n$, and $\varepsilon_n$ clear.

1. The directed $k$-nearest neighbor graph is defined by setting $h(x) = 1_{x \in [0,1]}$, the indicator function of the unit interval, $\varepsilon_n(x)$ the distance to the $k^{\text{th}}$ nearest neighbor, and $g_n = (k/n)^{1/d}$ the rate at which $\varepsilon_n(x)$ approaches zero.

2. A Gaussian kernel graph is approximated by setting $h(x) = \exp(-x^2/\sigma^2) 1_{x \in [0,1]}$. The truncation of the Gaussian tails at $\sigma$ is an analytic convenience rather than a fundamental limitation, and the bandwidth can be varied by rescaling $\varepsilon_n(x)$.

## 2.2 Continuum limit of the random walk

Our techniques rely on analysis of the limiting behavior of the simple random walk $X_t^n$ on a spatial graph $G_n$, viewed as a discrete-time Markov process with domain $D$. The increment at step $t$ of $X_t^n$ is a jump to a random point in $\mathcal{X}_n$ which lies within the ball of radius $\varepsilon_n(X_t^n)$ around $X_t^n$. We observe three effects: (A) the random walk jumps more frequently towards regions of high density; (B) the random walk moves more quickly whenever $\varepsilon_n(X_t^n)$ is large; (C) for $\varepsilon_n$ small and a large step count $t$, the random variable $X_t^n - X_0^n$ is the sum of many small independent (but not necessarily identically distributed) increments. In the $n \to \infty$ limit, we may identify $X_t^n$ with a continuous-time stochastic process satisfying (A), (B), and (C) via the following result, which is a slight strengthening of [6, Theorem 3.4] obtained by applying [15, Theorem 11.2.3] in place of the original result of Stroock-Varadhan.

**Theorem 2.2.** *The simple random walk $X_t^n$ converges uniformly in Skorokhod space $\mathsf{D}([0,\infty), \overline{D})$ after a time scaling $\hat{t} = tg_n^2$ to the Itô process $Y_{\hat{t}}$ valued in the space of continuous functions $\mathsf{C}([0,\infty), \overline{D})$ with reflecting boundary conditions on $D$ defined by*

$$dY_{\hat{t}} = \nabla \log(p(Y_{\hat{t}})) \overline{\varepsilon}(Y_{\hat{t}})^2/3 d\hat{t} + \overline{\varepsilon}(Y_{\hat{t}})/\sqrt{3} dW_{\hat{t}}. \tag{1}$$

Effects (A), (B), and (C) may be seen in the stochastic differential equation (1) as follows. The direction of the drift is controlled by $\nabla \log(p(Y_{\hat{t}}))$, the rate of drift is controlled by $\overline{\varepsilon}(Y_{\hat{t}})^2$, and the noise is driven by a Brownian motion $W_{\hat{t}}$ with location-dependent scaling $\overline{\varepsilon}(Y_{\hat{t}})/\sqrt{3}$.[1]

We view Theorem 2.2 as a method to understand the simple random walk $X_t^n$ through the continuous walk $Y_{\hat{t}}$. Attributes of stochastic processes such as stationary distribution or hitting time may be defined for both $Y_{\hat{t}}$ and $X_t^n$, and in many cases Theorem 2.2 implies that an appropriately-rescaled version of the discrete attribute will converge to the continuous one. Because attributes of the continuous process $Y_{\hat{t}}$ can reveal information about proximity between points, this provides a general framework for inference in spatial graphs. We use hitting times of the continuous process to a domain $E \subset D$ to prove properties of the hitting time of a simple random walk on a graph via the limit arguments of Theorem 2.2.

## 3 Degeneracy of expected hitting times in networks

The hitting time, commute time, and resistance distance are popular measures of distance based upon the random walk which are believed to be robust and capture the cluster structure of the network. However, it was shown in a surprising result in [19] that on undirected geometric graphs the scaled expected hitting time from $x_i$ to $x_j$ converges to inverse of the degree of $x_j$.

In Theorem 3.5, we give an intuitive explanation and generalization of this result by showing that if the random walk on a graph converges to any limiting Itô process in dimension $d \geq 2$, the scaled expected hitting time to any point converges to the inverse of the stationary distribution. This answers the open problem in [19] on the degeneracy of hitting times for directed graphs and graphs with general degree distributions such as directed $k$-nearest neighbor graphs, lattices, and power-law graphs with convergent random walks. Our proof can be understood as first extending the transience or neighborhood recurrence of Brownian motion for $d \geq 2$ to more general Itô processes and then connecting hitting times on graphs to their Itô process equivalents.

### 3.1 Typical hitting times are large

We will prove the following lemma that hitting a given vertex quickly is unlikely. Let $T_{x_j,n}^{x_i}$ be the hitting time to $x_j$ of $X_t^n$ started at $x_i$ and $T_E^{x_i}$ be the continuous equivalent for $Y_{\hat{t}}$ to hit $E \subset D$.

---

[1] Both the variance $\Theta(\varepsilon_n(x)^2)$ and expected value $\Theta(\nabla \log(p(x))\varepsilon_n(x)^2)$ of a single step in the simple random walk are $\Theta(g_n^2)$. The time scaling $\hat{t} = tg_n^2$ in Theorem 2.2 was chosen so that as $n \to \infty$ there are $g_n^{-2}$ discrete steps taken per unit time, meaning the total drift and variance per unit time tend to a non-trivial limit.

**Lemma 3.1** (Typical hitting times are large)**.** *For any $d \geq 2$, $c > 0$, and $\delta > 0$, for large enough $n$ we have $\mathbb{P}(T^{x_i}_{x_j,n} > cg_n^{-2}) > 1 - \delta$.*

To prove Lemma 3.1, we require the following tail bound following from the Feynman-Kac theorem.

**Theorem 3.2** ([10, Exercise 9.12] Feynman-Kac for the Laplace transform)**.** *The Laplace transform of the hitting time (LTHT) $u(x) = \mathbb{E}[\exp(-\beta T^x_E)]$ is the solution to the boundary value problem with boundary condition $u|_{\partial E} = 1$:*

$$\frac{1}{2} Tr[\sigma^T H(u)\sigma] + \mu(x) \cdot \nabla u - \beta u = 0.$$

This will allow us to bound the hitting time to the ball $B(x_j, s)$ of radius $s$ centered at $x_j$.

**Lemma 3.3.** *For $x, y \in D$, $d \geq 2$, and any $\delta > 0$, there exists $s > 0$ such that $\mathbb{E}[e^{-T^x_{B(y,s)}}] < \delta$.*

*Proof.* We compare the Laplace transformed hitting time of the general Itô process to that of Brownian motion via Feynman-Kac and handle the latter case directly. Details are in Section S2.1. □

We now use Lemma 3.3 to prove Lemma 3.1.

*Proof of Lemma 3.1.* Our proof proceeds in two steps. First, we have $T^{x_i}_{x_j,n} \geq T^{x_i}_{B(x_j,s),n}$ a.s. for any $s > 0$ because $x_j \in B(x_j, s)$, so by Theorem 2.2, we have

$$\lim_{n \to \infty} \mathbb{E}[e^{-T^{x_i}_{x_j,n} g_n^{-2}}] \leq \lim_{n \to \infty} \mathbb{E}[e^{-T^{x_i}_{B(x_j,s),n} g_n^{-2}}] = \mathbb{E}[e^{-T^{x_i}_{B(x_j,s)}}]. \tag{2}$$

Applying Lemma 3.3, we have $\mathbb{E}[e^{-T^{x_i}_{B(x_j,s)}}] < \frac{1}{2}\delta e^{-c}$ for some $s > 0$. For large enough $n$, this combined with (2) implies $\mathbb{P}(T^{x_i}_{x_j,n} \leq cg_n^{-2})e^{-c} < \delta e^{-c}$ and hence $\mathbb{P}(T^{x_i}_{x_j,n} \leq cg_n^{-2}) < \delta$. □

### 3.2 Expected hitting times degenerate to the stationary distribution

To translate results from Itô processes to directed graphs, we require a regularity condition. Let $q_t(x_j, x_i)$ denote the probability that $X^n_t = x_j$ conditioned on $X^n_0 = x_i$. We make the following technical conjecture which we assume holds for all spatial graphs.

$(\star)$ For $t = \Theta(g_n^{-2})$, the rescaled marginal $nq_t(x, x_i)$ is a.s. eventually uniformly equicontinuous.[2]

Let $\pi_{X^n}(x)$ denote the stationary distribution of $X^n_t$. The following was shown in [6, Theorem 2.1] under conditions implied by our condition $(\star)$ (Corollary S2.6).

**Theorem 3.4.** *Assuming $(\star)$, for $a^{-1} = \int p(x)^2 \overline{\varepsilon}(x)^{-2} dx$, we have the a.s. limit*

$$\widehat{\pi}(x) := \lim_{n \to \infty} n\pi_{X^n}(x) = a\frac{p(x)}{\overline{\varepsilon}(x)^2}.$$

We may now express the limit of expected hitting time in terms of this result.

**Theorem 3.5.** *For $d \geq 2$ and any $i, j$, we have*

$$\frac{\mathbb{E}[T^{x_i}_{x_j,n}]}{n} \xrightarrow{a.s.} \frac{1}{\widehat{\pi}(x_j)}.$$

*Proof.* We give a sketch. By Lemma 3.1, the random walk started at $x_i$ does not hit $x_j$ within $cg_n^{-2}$ steps with high probability. By Theorem S2.5, the simple random walk $X^n_t$ mixes at exponential rate, implying in Lemma S2.8 that the probability of first hitting at step $t > cg_n^{-2}$ is approximately the stationary distribution at $x_j$. Expected hitting time is then shown to approximate the expectation of a geometric random variable. See Section S2 for a full proof. □

Theorem 3.5 is illustrated in Figures 1A and 1B, which show with only 3000 points, expected hitting times on a $k$-nearest neighbor graph degenerate to the stationary distribution.[3]

---

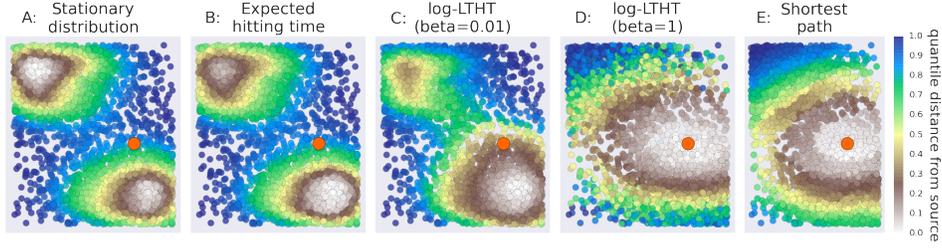

Figure 1: Estimated distance from orange starting point on a $k$-nearest neighbor graph constructed on two clusters. A and B show degeneracy of hitting times (Theorem 3.5). C, D, and E show that log-LTHT interpolate between hitting time and shortest path.

## 4 The Laplace transformed hitting time (LTHT)

In Theorem 3.5 we showed that expected hitting time is degenerate because a simple random walk mixes before hitting its target. To correct this we penalize longer paths. More precisely, consider for $\widehat{\beta} > 0$ and $\beta_n = \widehat{\beta} g_n^2$ the *Laplace transforms* $\mathbb{E}[e^{-\widehat{\beta} T_E^x}]$ and $\mathbb{E}[e^{-\beta_n T_{E,n}^x}]$ of $T_E^x$ and $T_{E,n}^x$.

These Laplace transformed hitting times (LTHT's) have three advantages. First, while the expected hitting time of a Brownian motion to a domain is dominated by long paths, the LTHT is dominated by direct paths. Second, the LTHT for the Itô process can be derived in closed form via the Feynman-Kac theorem, allowing us to make use of techniques from continuous stochastic processes to control the continuum LTHT. Lastly, the LTHT can be computed both by sampling and in closed form as a matrix inversion (Section S3). Now define the scaled log-LTHT as

$$-\log(\mathbb{E}[e^{-\beta_n T_{x_j,n}^{x_i}}])/\sqrt{2\beta_n} g_n.$$

Taking different scalings for $\beta_n$ with $n$ interpolates between expected hitting time ($\beta_n \to 0$ on a fixed graph) and shortest path distance ($\beta_n \to \infty$) (Figures 1C, D, and E). In Theorem 4.5, we show that the intermediate scaling $\beta_n = \Theta(\widehat{\beta} g_n^2)$ yields a consistent distance measure retaining the unique properties of hitting times. Most of our results on the LTHT are novel for any quasi-walk metric.

While considering the Laplace transform of the hitting time is novel to our work, this metric has been used in the literature in an ad-hoc manner in various forms as a similarity metric for collaboration networks [20], hidden subgraph detection [14], and robust shortest path distance [21]. However, these papers only considered the elementary properties of the limits $\beta_n \to 0$ and $\beta_n \to \infty$. Our consistency proof demonstrates the advantage of the stochastic process approach.

### 4.1 Consistency

It was shown previously that for $n$ fixed and $\beta_n \to \infty$, $-\log(\mathbb{E}[-\beta_n T_{x_j,n}^{x_i}])/\beta_n g_n$ converges to shortest path distance from $x_i$ to $x_j$. We investigate more precise behavior in terms of the scaling of $\beta_n$. There are two regimes: if $\beta_n = \omega(\log(g_n^d n))$, then the shortest path dominates and the LTHT converges to shortest path distance (See Theorem S5.2). If $\beta_n = \Theta(\widehat{\beta} g_n^2)$, the graph log-LTHT converges to its continuous equivalent, which for large $\widehat{\beta}$ averages over random walks concentrated around the geodesic. To show consistency for $\beta_n = \Theta(\widehat{\beta} g_n^2)$, we proceed in three steps: (1) we reweight the random walk on the graph so the limiting process is Brownian motion; (2) we show that log-LTHT for Brownian motion recovers latent distance; (3) we show that log-LTHT for the reweighted walk converges to its continuous limit; (4) we conclude that log-LTHT of the reweighted walk recovers latent distance.

**(1) Reweighting the random walk to converge to Brownian motion:** We define weights using the estimators $\widehat{p}$ and $\widehat{\varepsilon}$ for $p(x)$ and $\overline{\varepsilon}(x)$ from [6].

---

times to small out neighbors which corrects this problem and derive closed form solutions (Theorem S2.12). This hitting time is non-degenerate but highly biased due to boundary terms (Corollary S2.14).

**Theorem 4.1.** *Let $\widehat{p}$ and $\widehat{\varepsilon}$ be consistent estimators of the density and local scale and $A$ be the adjacency matrix. Then the random walk $\widehat{X}^n_t$ defined below converges to a Brownian motion.*

$$\mathbb{P}(\widehat{X}^n_{t+1} = x_j \mid \widehat{X}^n_t = x_i) = \begin{cases} \frac{A_{i,j}\widehat{p}(x_j)^{-1}}{\sum_k A_{i,k}\widehat{p}(x_k)^{-1}}\widehat{\varepsilon}(x_i)^{-2} & i \neq j \\ 1 - \widehat{\varepsilon}(x_i)^{-2} & i = j \end{cases}$$

*Proof.* Reweighting by $\widehat{p}$ and $\widehat{\varepsilon}$ is designed to cancel the drift and diffusion terms in Theorem 2.2 by ensuring that as $n$ grows large, jumps have means approaching $0$ and variances which are asymptotically equal (but decaying with $n$). See Theorem S4.1. [4] □

**(2) Log-LTHT for a Brownian motion:** Let $W_t$ be a Brownian motion with $W_0 = x_i$, and let $\overline{T}^{x_i}_{B(x_j,s)}$ be the hitting time of $W_t$ to $B(x_j, s)$. We show that log-LTHT converges to distance.

**Lemma 4.2.** *For any $\alpha < 0$, if $\widehat{\beta} = s^\alpha$, as $s \to 0$ we have*

$$-\log(\mathbb{E}[\exp(-\widehat{\beta}\overline{T}^{x_i}_{B(x_j,s)})])/\sqrt{2\widehat{\beta}} \to |x_i - x_j|.$$

*Proof.* We consider hitting time of Brownian motion started at distance $|x_i - x_j|$ from the origin to distance $s$ of the origin, which is controlled by a Bessel process. See Subsection S6.1 for details. □

**(3) Convergence of LTHT for $\beta_n = \Theta(\widehat{\beta}g_n^2)$:** To compare continuous and discrete log-LTHT's, we will first define the $s$-neighborhood of a vertex $x_i$ on $G_n$ as the graph equivalent of the ball $B(x_i, s)$.

**Definition 4.3** ($s$-neighborhood). *Let $\widehat{\varepsilon}(x)$ be the consistent estimate of the local scale from [6] so that $\widehat{\varepsilon}(x) \to \overline{\varepsilon}(x)$ uniformly a.s. as $n \to \infty$. The $\widehat{\varepsilon}$-weight of a path $x_{i_1} \to \cdots \to x_{i_l}$ is the sum $\sum_{m=1}^{l-1} \widehat{\varepsilon}(x_{i_m})$ of vertex weights $\widehat{\varepsilon}(x_i)$. For $s > 0$ and $x \in G_n$, the $s$-neighborhood of $x$ is*

$$\mathsf{NB}^s_n(x) := \{y \mid \text{there is a path } x \to y \text{ of } \widehat{\varepsilon}\text{-weight} \leq g_n^{-1}s\}.$$

For $x_i, x_j \in G_n$, let $\widehat{T}^{x_i}_{B(x_j,s)}$ be the hitting time of the transformed walk on $G_n$ from $x_i$ to $\mathsf{NB}^s_n(x_j)$. We now verify that hitting times to the $s$-neighborhood on graphs and the $s$-radius ball coincide.

**Corollary 4.4.** *For $s > 0$, we have $g_n^2\widehat{T}^{x_i}_{\mathsf{NB}^s_n(x_j),n} \xrightarrow{d} \overline{T}^{x_i}_{B(x_j,s)}$.*

*Proof.* We verify that the ball and the neighborhood have nearly identical sets of points and apply Theorem 2.2. See Subsection S6.2 for details. □

**(4) Proving consistency of log-LTHT:** Properly accounting for boundary effects, we obtain a consistency result for the log-LTHT for small neighborhood hitting times.

**Theorem 4.5.** *Let $x_i, x_j \in G_n$ be connected by a geodesic not intersecting $\partial D$. For any $\delta > 0$, there exists a choice of $\widehat{\beta}$ and $s > 0$ so that if $\beta_n = \widehat{\beta}g_n^2$, for large $n$ we have with high probability*

$$\left| -\log(\mathbb{E}[\exp(-\beta_n\widehat{T}^{x_i}_{\mathsf{NB}^s_n(x_j),n})])/\sqrt{2\widehat{\beta}} - |x_i - x_j| \right| < \delta.$$

*Proof of Theorem 4.5.* The proof has three steps. First, we convert to the continuous setting via Corollary 4.4. Second, we show the contribution of the boundary is negligible. The conclusion follows from the explicit computation of Lemma S6.1. Full details are in Section S6. □

The stochastic process limit based proof of Theorem 4.5 implies that the log-LTHT is consistent and robust to small perturbations to the graph which preserve the same limit (Supp. Section S8).

---

[4]This is a special case of a more general theorem for transforming limits of graph random walks (Theorem S4.1). Figure S1 shows that this modification is highly effective in practice.

### 4.2 Bias

Random walk based metrics are often motivated as recovering a cluster preserving metric. We now show that the log-LTHT of the un-weighted simple random walk preserves the underlying cluster structure. In the 1-D case, we provide a complete characterization.

**Theorem 4.6.** *Suppose the spatial graph has $d = 1$ and $h(x) = 1_{x \in [0,1]}$. Let $T^{x_i}_{\mathsf{NB}_n^{\bar{\varepsilon}(x_j)g_n}(x_j),n}$ be the hitting time of a simple random walk from $x_i$ to the out-neighborhood of $x_j$. It converges to*

$$-\log(\mathbb{E}[-\beta T^{x_i}_{\mathsf{NB}_n^{\bar{\varepsilon}(x_j)g_n}(x_j),n}])/\sqrt{8\beta} \to \int_{x_i}^{x_j} \sqrt{m(x)}dx + o\left(\log(1 + e^{-\sqrt{2\beta}})/\sqrt{2\beta}\right),$$

*where $m(x) = \frac{2}{\bar{\varepsilon}(x)^2} + \frac{1}{\beta}\frac{\partial \log(p(x))}{\partial x^2} + \frac{1}{\beta}\left(\frac{\partial \log(p(x))}{\partial x}\right)^2$ defines a density-sensitive metric.*

*Proof.* Apply the WKBJ approximation for Schrodinger equations to the Feynman-Kac PDE from Theorem 3.2. See Corollary S7.2 and Corollary S2.13 for a full proof. □

The leading order terms of the density-sensitive metric appropriately penalize crossing regions of large changes to the log density; this is not the case for the expected hitting time (Theorem S2.12).

### 4.3 Robustness

While shortest path distance is a consistent measure of the underlying metric, it breaks down catastrophically with the addition of a single non-geometric edge and does not meaningfully rank vertices that share an edge. In contrast, we show that LTHT breaks ties between vertices via the resource allocation (RA) index, a robust local similarity metric under Erdős-Rényi-type noise. [5]

**Definition 4.7.** *The noisy spatial graph $G_n$ over $\mathcal{X}_n$ with noise terms $q_1(n)$, ..., $q_n(n)$ is constructed by drawing an edge from $x_i$ to $x_j$ with probability*

$$p_{ij} = h(|x_i - x_j|\varepsilon_n(x_i)^{-1})(1 - q_j(n)) + q_j(n).$$

Define the directed RA index in terms of the out-neighborhood set $\mathsf{NB}_n(x_i)$ and the in-neighborhood set $\mathsf{NB}_n^{\text{in}}(x_i)$ as $R_{ij} := \sum_{x_k \in \mathsf{NB}_n(x_i) \cap \mathsf{NB}_n^{\text{in}}(x_j)} |\mathsf{NB}_n(x_k)|^{-1}$ and two step log-LTHT by $M_{ij}^{\text{ts}} := -\log(\mathbb{E}[\exp(-\beta T^{x_i}_{x_j,n}) \mid T^{x_i}_{x_j,n} > 1])$. [6] We show two step log-LTHT and RA index give equivalent methods for testing if vertices are within distance $\varepsilon_n(x)$.

**Theorem 4.8.** *If $\beta = \omega(\log(g_n^d n))$ and $x_i$ and $x_j$ have at least one common neighbor, then*

$$M_{ij}^{\text{ts}} - 2\beta \to -\log(R_{ij}) + \log(|\mathsf{NB}_n(x_i)|).$$

*Proof.* Let $P_{ij}(t)$ be the probability of going from $x_i$ to $x_j$ in $t$ steps, and $H_{ij}(t)$ the probability of not hitting before time $t$. Factoring the two-step hitting time yields

$$M_{ij}^{\text{ts}} = 2\beta - \log(P_{ij}(2)) - \log\left(1 + \sum_{t=3}^{\infty} \frac{P_{ij}(t)}{P_{ij}(2)} H_{ij}(t)e^{-\beta(t-2)}\right).$$

Let $k_{\max}$ be the maximal out-degree in $G_n$. The contribution of paths of length greater than 2 vanishes because $H_{ij}(t) \leq 1$ and $P_{ij}(t)/P_{ij}(2) \leq k_{\max}^2$, which is dominated by $e^{-\beta}$ for $\beta = \omega(\log(g^n n))$. Noting that $P_{ij}(2) = \frac{R_{ij}}{|\mathsf{NB}_n(x_i)|}$ concludes. For full details see Theorem S9.1. □

For edge identification within distance $\varepsilon_n(x)$, the RA index is robust even at noise level $q = o(g_n^{d/2})$.

---

[5] Modifying the graph by changing fewer than $g_n^2/n$ edges does not affect the continuum limit of the random graph, and therefore preserve the LTHT with parameter $\beta = \Theta(g_n^2)$. While this weak bound allows on average $o(1)$ noise edges per vertex, it does show that the LTHT is substantially more robust than shortest paths without modification. See Section S8 for proofs.

[6] The conditioning $T^{x_i}_{x_j,n} > 1$ is natural in link-prediction tasks where only pairs of disconnected vertices are queried. Empirically, we observe it is critical to performance (Figure 3).

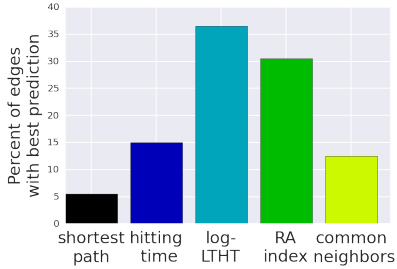

Figure 2: The LTHT recovered deleted edges most consistently on a citation network

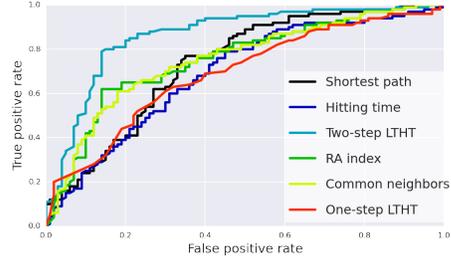

Figure 3: The two-step LTHT (defined above Theorem 4.8) outperforms others at word similarity estimation including the basic log-LTHT.

**Theorem 4.9.** *If $q_i = q = o(g_n^{d/2})$ for all $i$, for any $\delta > 0$ there are $c_1, c_2$ and $h_n$ so that for any $i, j$, with probability at least $1 - \delta$ we have*

- $|x_i - x_j| < \min\{\varepsilon_n(x_i), \varepsilon_n(x_j)\}$ *if $R_{ij}h_n < c_1$;*
- $|x_i - x_j| > 2\max\{\varepsilon_n(x_i), \varepsilon_n(x_j)\}$ *if $R_{ij}h_n > c_2$.*

*Proof.* The minimal RA index follows from standard concentration arguments (see S9.2). □

## 5 Link prediction tasks

We compare the LTHT against other baseline measures of vertex similarity: shortest path distance, expected hitting time, number of common neighbors, and the RA index. A comprehensive evaluation of these quasi-walk metrics was performed in [8] who showed that a metric equivalent to the LTHT performed best. We consider two separate link prediction tasks on the largest connected component of vertices of degree at least five, fixing $\beta = 0.2$.[7] The degree constraint is to ensure that local methods using number of common neighbors such as the resource allocation index do not have an excessive number of ties. Code to generate figures in this paper are contained in the supplement.

**Citation network:** The KDD 2003 challenge dataset [5] includes a directed, unweighted network of e-print arXiv citations whose dense connected component has 11,042 vertices and 222,027 edges. We use the same benchmark method as [9] where we delete a single edge and compare the similarity of the deleted edge against the set of control pair of vertices $i, j$ which do not share an edge. We count the fraction of pairs on which each method rank the deleted edge higher than all other methods. We find that LTHT is consistently best at this task (Figure 2). [8]

**Associative Thesaurus network:** The Edinburgh associative thesaurus [7] is a network with a dense connected component of 7754 vertices and 246,609 edges in which subjects were shown a set of ten words and for each word was asked to respond with the first word to occur to them. Each vertex represents a word and each edge is a weighted, directed edge where the weight from $x_i$ to $x_j$ is the number of subjects who responded with word $x_j$ given word $x_i$.

We measure performance by whether strong associations with more than ten responses can be distinguished from weak ones with only one response. We find that the LTHT performs best and that preventing one-step jumps is critical to performance as predicted by Theorem 4.8 (Figure 3).

## 6 Conclusion

Our work has developed an asymptotic equivalence between hitting times for random walks on graphs and those for diffusion processes. Using this, we have provided a short extension of the proof for the divergence of expected hitting times, and derived a new consistent graph metric that is theoretically principled, computationally tractable, and empirically successful at well-established link prediction benchmarks. These results open the way for the development of other principled quasi-walk metrics that can provably recover underlying latent similarities for spatial graphs.

---

[7] Results are qualitatively identical when varying $\beta$ from 0.1 to 1; see supplement for details.

[8] The two-step LTHT is not shown since it is equivalent to the LTHT in missing link prediction.

# Supplementary materials and proofs

October 30, 2015

## Contents

## 1   Uniform equicontinuity of the marginals

We discuss condition ($\star$) stated in the main text. We conjecture and assume that the following technical condition holds on all spatial graphs.

($\star$) For $t = \Theta(g_n^{-2})$, the rescaled marginal density $nq_t(x, x_i)$ is a.s. eventually uniformly equicontinuous.

To make the terminology we use in ($\star$) clear, we rephrase it as follows.

**Definition S1.1** (Condition $\star$). *If $t = \Theta(g_n^{-2})$, with probability 1, for any $\delta > 0$, there exist $\gamma > 0$ and $n_0$ so that for $n > n_0$, any $x_k \in \mathcal{X}_n$, and any $x_i, x_j \in \mathcal{X}_n$ with $|x_i - x_j| < \gamma$, we have*

$$|nq_t(x_i, x_k) - nq_t(x_j, x_k)| < \delta.$$

This statement allows us to convert the weak convergence in distribution ensured by Theorem 2.2 and the results of (Stroock & Varadhan, 1971) to the convergence in density required by Corollary S2.6. Such a statement rules out the possibility that the density function $q_t(x, x_i)$ oscillates at frequencies increasing with $n$ as $n \to \infty$. Controlling regularity of $q_t(x, x_i)$ seems to be a critical ingredient for approaching $(\star)$.

In the case of an undirected graph, $(\star)$ follows from results in the literature. The strong local limit law for simple random walks shown in (Croydon & Hambly, 2008) yields an even stronger result than $(\star)$. In addition, in the same setting, the convergence result for spectral clustering of (von Luxburg et al., 2008) yields an equicontinuity result for eigenvectors of the Laplacian which implies $(\star)$.

However, for directed graphs no such result exists, and non-reversibility of the Markov chain seems to be an obstacle to proving such a result. Some results on utilizing the directed Laplacian as a smoothing operator (Zhou et al., 2005) exist in this direction. We believe these techniques could lead to an approach to $(\star)$ but thus far they have not yielded a sufficiently strong equicontinuity result.

*Remark* 1. In Hashimoto et al. (2015), a similar conjecture was made on the uniform equicontinuity of the rescaled stationary distribution $n\pi_{\mathcal{X}^n}(x)$. We believe "uniform equicontinuity" should be corrected to "eventual uniform equicontinuity" there. In Corollary S2.6, we show that $(\star)$ implies the conjecture of Hashimoto et al. (2015).

One consequence of equicontinuity is that convergence in distribution implies convergence in density. We prove this for the marginal distribution $\widehat{q_t}(x_k, x_i)$ of $Y_{\widehat{t}}$ for the purpose of Theorem 3.5, following the original strategy of (Hashimoto et al., 2015).

**Lemma S1.2** (Convergence of marginal densities). *If $t_n g_n^2 = \widehat{t} = \Theta(1)$, then under $(\star)$ we have*

$$\lim_{n \to \infty} n q_{t_n}(x, x_i) = \frac{\widehat{q_{\widehat{t}}}(x, x_i)}{p(x)},$$

*where the convergence is uniform in $x$ and $x_i$.*

*Proof.* The a.s. weak convergence of processes of Theorem 2.2 (which is uniform in $x_i$) implies by (Ethier & Kurtz, 1986, Theorem 4.9.12) that the empirical marginal distribution

$$d\mu_n = \sum_{j=1}^{n} q_{t_n}(x_j, x_i) \delta_{x_j}$$

converges weakly to the marginal distribution $d\mu = \widehat{q_{\widehat{t}}}(x, x_i) dx$ for $Y_{\widehat{t}}$. For any $x \in \mathcal{X}$ and $\delta > 0$, weak convergence against the test function $1_{B(x,\delta)}$ yields

$$\sum_{y \in \mathcal{X}_n, |y-x|<\delta} q_{t_n}(y, x_i) \to \int_{|y-x|<\delta} \widehat{q_{\widehat{t}}}(y, x_i) dy.$$

By eventual uniform equicontinuity of $n q_t(x, x_i)$, for any $\varepsilon > 0$ there is small enough $\delta > 0$ so that for all $n$ we have

$$\left| \sum_{y \in \mathcal{X}_n, |y-x|<\delta} q_{t_n}(y, x_i) - |\mathcal{X}_n \cap B(x, \delta)| q_{t_n}(x, x_i) \right| \leq n^{-1} |\mathcal{X}_n \cap B(x, \delta)| \varepsilon,$$

which implies that

$$\lim_{n \to \infty} q_{t_n}(x, x_i) p(x) n = \lim_{\delta \to 0} \lim_{n \to \infty} V_d^{-1} \delta^{-d} n q_{t_n}(x, x_i) \int_{|y-x|<\delta} p(y) dy$$

$$= \lim_{\delta \to 0} \lim_{n \to \infty} V_d^{-1} \delta^{-d} |\mathcal{X}_n \cap B(x, \delta)| q_{t_n}(x, x_i) = \lim_{\delta \to 0} V_d^{-1} \delta^{-d} \int_{|y-x|<\delta} \widehat{q_{\widehat{t}}}(y, x_i) dy = \widehat{q_{\widehat{t}}}(x, x_i).$$

We conclude the desired

$$\lim_{n \to \infty} n q_t(x, x_i) = \frac{\widehat{q_{\widehat{t}}}(x, x_i)}{p(x)},$$

where uniformity in $x$ comes from $(\star)$ and uniformity in $x_i$ comes from uniformity of Theorem 2.2. $\qquad \square$

## 2 Hitting times

In this section, we prove Theorem 3.5 to generalize the result of von Luxburg et al. (2014) on degenerate behavior of hitting times via Lemma 3.1. Our proof consists of two parts. First, we show that by Lemma 3.1 we can make the random walk mix before hitting any point. Next, we use this to show that if the chain is sufficiently mixed, then the expected hitting time is degenerate.

### 2.1 Typical hitting times are large

In this subsection, we give a complete proof of Lemma 3.3, reproduced here. Recall that $T_E^{x_i}$ is the hitting time of $Y_t$ from $x_i$ to a domain $E \subset D$. We will require a more general version of the Feynman-Kac theorem.

**Theorem S2.1** ((Øksendal, 2003, Exercise 9.12) Feynman-Kac). *Let $Z_t$ be an Itô process in $\mathbb{R}^d$ defined by*

$$dZ_t = \mu(Z_t)dt + \sigma(Z_t)dB_t.$$

*For a function $V(x)$ and $T_E^x$ the hitting time to a domain $E \subset D$, the function*

$$u(x) = \mathbb{E}\left[e^{-\int_0^{T_E^x} V(Z_s)ds}\right]$$

*is the solution to the boundary value problem*

$$\frac{1}{2}Tr[\sigma^T Hu\sigma] + \mu(x) \cdot \nabla u - V(x)u = 0$$

*with boundary condition $u|_{\partial E} = 1$.*

**Lemma S2.2.** *For $x, y \in D$, $d \geq 2$, and any $\delta > 0$, there exists $s > 0$ such that $\mathbb{E}[e^{-T_{B(y,s)}^x}] < \delta$.*

*Proof.* We use Feynman-Kac to compare $\mathbb{E}[e^{-T_{B(y,s)}^x}]$ for the general process to that of Brownian motion. By Theorem 3.2, $u_s(x) = \mathbb{E}[e^{-T_{B(y,s)}^x}]$ satisfies the boundary value problem

$$\Delta u_s + 2\nabla p(x) \cdot \nabla u_s - 2u_s \bar{\varepsilon}(x)^{-2} = 0$$

with $u_s|_{B(y,s)} \equiv 1$. This is equivalent to

$$\sum_i \left(\partial_i[p(x)^2 \partial_i u_s] - \frac{2}{d}u_s \frac{p(x)^2}{\bar{\varepsilon}(x)^2}\right) = 0.$$

Set $v_s(x) = p(x)u_s(x)$ and change variables to obtain

$$\sum_i \left(\partial_i[p(x)\partial_i v_s - v_s(x)\partial_i p(x)] - \frac{2}{d}v_s \frac{p(x)}{\bar{\varepsilon}(x)^2}\right) = p(x)\Delta v_s - \Delta p(x)v_s - v_s \frac{2p(x)}{\bar{\varepsilon}(x)^2} = 0,$$

which is equivalent to

$$\Delta v_s - \left(\frac{\Delta p(x)}{p(x)} + 2\bar{\varepsilon}(x)^{-2}\right)v_s = 0$$

with boundary condition $v_s|_{\partial B(y,s)} = 1$. Theorem S2.1 for $V(x) = \frac{\Delta p(x)}{p(x)} + 2\bar{\varepsilon}(x)^{-2}$ implies

$$v_s(x) = \mathbb{E}\left[e^{-\int_0^{\overline{T}_{B(y,s)}^x} \frac{\Delta p(B_r)}{p(B_r)} + 2\bar{\varepsilon}(B_r)^{-2}dr}\right]$$

for $B_t$ Brownian motion started at $x$ and $\overline{T}_{B(y,s)}^x$ the hitting time of $B_t$ to $B(y,s)$. For a constant $C$ depending on $p$ and $\bar{\varepsilon}$, we have

$$u_s(x) = \frac{v_s(x)}{p(x)} \leq \mathbb{E}[e^{-C\overline{T}_{B(y,s)}^x}].$$

Applying Lemma S2.3 with this $C$ implies $u_s(x) < \delta e^{-c}$, as needed. □

**Lemma S2.3.** *For $x, y \in D$, let $B_t$ be a Brownian motion with reflecting boundary condition in $D$ started at $x$ and $T^x_{B(y,s)}$ its hitting time to $B(y,s)$. Then for any sufficiently small $C, c, \delta > 0$, there exists some $s > 0$ so that*

$$\mathbb{E}[e^{-CT_{B(y,s)}}] < \delta e^{-c}.$$

*Proof.* Fix $c > 0$ and $\delta > 0$ small enough so that $e^{-c}\delta < 1/10$. If $|x - y| = p$, by Theorem 3 of Byczkowski et al. (2013), the probability density of $T^x_{B(y,s)}$ started at $x$ if there were no outer boundary is bounded by

$$p(t) < C_1 \frac{s^3(p-s)e^{-\frac{(p-s)^2}{2t}}}{pt^{3/2}} \begin{cases} ((t/s^2)^{\frac{d-3}{2}} + (p/s)^{\frac{d-3}{2}})^{-1} & d > 2 \\ \frac{(p/s+t/s^2)^{1/2}(1+\log(p/s))}{(1+\log(1+t/ps))(1+\log(p/s+t/s^2))} & d = 2 \end{cases}$$

for some constant $C_1$.

**Choosing constants:** We claim that we can choose $s$, $p$, and $r$ with $r > p > s > 0$ so that:

(a) $B(y, r)$ is contained entirely in the domain $D$;

(b1) $\frac{r^{2-d}-p^{2-d}}{r^{2-d}-s^{2-d}} > \max\{1/2, \frac{1-e^{-c}\delta}{1-\frac{1}{2}e^{-c}\delta}\}$ if $d > 2$;

(b2) $\frac{\log r - \log p}{\log r - \log s} > \max\{1/2, \frac{1-e^{-c}\delta}{1-\frac{1}{2}e^{-c}\delta}\}$ if $d = 2$;

(c1) $C_1 s^d \left( C^{-1} + \int_1^\infty u^{d/2-2}e^{-(p-s)^2 u}du \right) < \frac{1}{4}\delta e^{-c}$ if $d > 2$;

(c2a) $C_1 s^2 \int_1^\infty e^{-Ct}(ps+t)^{1/2}dt < \frac{1}{8}\delta e^{-c}$ if $d = 2$;

(c2b) $C_1 s^2(ps+1)^{1/2} \int_0^1 t^{-3/2}e^{-(p-s)^2/2t}dt < \frac{1}{8}\delta e^{-c}$ if $d = 2$.

For $d > 2$, we have that

$$\frac{\Gamma(d/2-1)}{(p-s)^{d-2}} = (p-s)^{2-d}\int_0^\infty x^{d/2-2}e^{-x}dx = \int_0^\infty u^{d/2-2}e^{-(p-s)^2 u}du > \int_1^\infty u^{d/2-2}e^{-(p-s)^2 u}du.$$

Then for $p = 2qr$ and $s = qr$, we have that

$$\frac{r^{2-d}-p^{2-d}}{r^{2-d}-s^{2-d}} = \frac{1-2^{d-2}q^{d-2}}{1-q^{d-2}} > 1 - 2^{d-2}q^{d-2}$$

and that

$$s^d \frac{\Gamma(d/2-1)}{(p-s)^{d-2}} < q^2 r^2 \Gamma(d/2-1)$$

Therefore, sending $r \to 0$ and $q \to 0$ gives a choice of $r > p > s$ satisfying (a), (b1), and (c1) as needed.

For $d = 2$, notice that for $t = u^{-1}$, we have

$$\int_0^1 t^{-3/2}e^{-(p-s)^2/2t}dt = \int_1^\infty u^{-1/2}e^{-(p-s)^2 u/2}du.$$

Observe now that

$$(p-s)^{-1}\Gamma(1/2) = (p-s)^{-1}\int_0^\infty t^{-1/2}e^{-t}dt = \int_0^\infty u^{-1/2}e^{-(p-s)^2 u}du > \int_1^\infty u^{-1/2}e^{-(p-s)^2 u}du,$$

whence we conclude that

$$C_1 \frac{s^2\sqrt{ps+1}}{p-s}\Gamma(1/2) > C_1 s^2(ps+1)^{1/2}\int_0^1 t^{-3/2}e^{-(p-s)^2/2t}dt.$$

Again choose $p = 2qr$ and $s = qr$, for which we obtain

$$C_1 s^2(ps+1)^{1/2}\int_0^1 t^{-3/2}e^{-(p-s)^2/2t}dt < C_1\Gamma(1/2)qr\sqrt{4q^2 r^2+1}.$$

Sending $q$ and $r$ to 0 then yields $r > p > s$ satisfying (a), (b2) because $q \to 0$, (c2a) because $s \to 0$ and $(ps + t)^{1/2} < (1 + t)^{1/2}$, and (c2b) by the estimate above.

**Bounding the Laplace transform:** Having chosen $r > p > s > 0$ with the desired properties, we have for any $z \in D$ that

$$\mathbb{E}[e^{-CT^z_{B(y,s)}}] \leq \max_{|x-y|=p} \mathbb{E}[e^{-CT^x_{B(y,s)}}].$$

Our strategy will be to bound $\mathbb{E}[e^{-CT^x_{B(y,s)}}]$ for any $x$ with $|x - y| = p$. Fix such an $x$. Let $E$ be the event that the walk hits $B(y, s)$ before $B(y, r)$. By Theorem 3.17 of Mörters & Peres (2010), the probability of $E$ is $\frac{r^{2-d}-p^{2-d}}{r^{2-d}-s^{2-d}}$ if $d > 2$ and $\frac{\log r - \log p}{\log r - \log s}$ if $d = 2$. By our choice of parameters, this probability is at least $\mathbb{P}(E) > \max\{1/2, \frac{1-e^{-c}\delta}{1-\frac{1}{2}e^{-c}\delta}\}$.

Let $\mathbb{E}'[e^{-CT^x_{B(y,s)}}]$ denote the case where there is no outside boundary. For $d > 2$, we have

$$\mathbb{E}'[e^{-CT^x_{B(y,s)}}] < C_1 s^d \int_0^\infty e^{-Ct} t^{-d/2} e^{-(p-s)^2/2t} dt$$

$$< C_1 s^d \left( C^{-1} + \int_0^1 t^{-d/2} e^{-(p-s)^2/2t} dt \right)$$

$$< C_1 s^d \left( C^{-1} + \int_1^\infty u^{d/2-2} e^{-(p-s)^2 u} du \right)$$

$$< \frac{1}{4} \delta e^{-c}$$

by the choice of $s$ and $p$. For $d = 2$, we have

$$\mathbb{E}'[e^{-CT^x_{B(y,s)}}] < C_1 s^2 \int_0^\infty e^{-Ct} t^{-3/2} e^{-(p-s)^2/2t} (ps + t)^{1/2} dt < \frac{1}{4} \delta e^{-c}$$

again by our choice of $s$ and $p$. Conditioning on $E$, we find that

$$\mathbb{E}[e^{-CT^x_{B(y,s)}} | E] \leq \mathbb{P}(E)^{-1} \mathbb{E}'[e^{-CT^x_{B(y,s)}}] < \frac{1}{2} \delta e^{-c}.$$

This implies the desired

$$\mathbb{E}[e^{-CT^x_{B(y,s)}}] \leq \mathbb{P}(E) \mathbb{E}[e^{-CT^x_{B(y,s)}}] + (1 - \mathbb{P}(E)) < \delta e^{-c}. \qquad \square$$

## 2.2 Exponential mixing on spatial graphs

In this subsection, we show that mixing rates are exponential on spatial graphs as assuming ($\star$).

**Lemma S2.4** (Uniform Doeblin condition). *Assuming ($\star$), there exist $\alpha > 0$ and $K < \infty$ so that for some $n_0 > 0$ and $\hat{t}_0 > 0$, we have for $n > n_0$ and $\hat{t} > \hat{t}_0$ that*

1. $\min_{x,x_i \in \mathcal{X}_n} q^n_{\lceil \hat{t} g_n^{-2} \rceil}(x, x_i) > \frac{\alpha}{n}$;

2. $\max_{x,x_i \in \mathcal{X}_n} q^n_{\lceil \hat{t} g_n^{-2} \rceil}(x, x_i) < \frac{K}{n}$.

*Proof.* By Lemma S1.2, assuming ($\star$) we have $n q^n_{\lceil \hat{t} g_n^{-2} \rceil}(x, x_i) \to \hat{q}_{\hat{t}}(x, x_i)/p(x)$, where convergence is uniform in $x$ and $x_i$. Therefore, we may choose $n_0 > 0$ and $\hat{t}_0 > 0$ so that for $n > n_0$ and $\hat{t} > \hat{t}_0$, we have for any $x, x_i \in \mathcal{X}_n$ that

$$\min_{x,x_i \in D} \hat{q}_{\hat{t}}(x, x_i)/p(x) \leq n q^n_{\lceil \hat{t} g_n^{-2} \rceil}(x, x_i) \leq \max_{x,x_i \in D} \hat{q}_{\hat{t}}(x, x_i)/p(x),$$

where the first and last quantities are well-defined by compactness of $D$. Taking

$$\alpha = \frac{1}{2} \min_{x,y \in D} \hat{q}_{\hat{t}}(x, y)/p(x) \qquad \text{and} \qquad K = 2 \max_{x,y \in D} \hat{q}_{\hat{t}}(x, y)/p(x)$$

thus fulfills the desired conditions. $\qquad \square$

**Theorem S2.5.** *Then we may choose $\hat{t}_0, n_0 > 0$ and $C, \beta > 0$ so that for $\hat{t} > \hat{t}_0$, $n > n_0$ and $x_i, x \in \mathcal{X}_n$, we have*

$$|q^n_{\lceil \hat{t} g_n^{-2} \rceil}(x, x_i) - \pi_{X_n}(x)| < C \exp(-\beta \hat{t}) \pi_{X_n}(x).$$

*Proof.* By Lemma S2.4, the family of processes $X^n_t$ satisfies the uniform Doeblin condition of (Eloranta, 1990, Section 2.8). The claim follows by the consequences for exponential mixing given in the analogue of (Eloranta, 1990, Theorem 2.7). □

**Corollary S2.6.** *Assuming ($\star$), the rescaled stationary distribution $n\pi_{X^n}(x)$ is a.s. eventually uniformly equicontinuous.*

*Proof.* Choose $\alpha, K, \hat{t}_1, n_1$ by Lemma S2.4 so that for $n > n_1, \hat{t} > \hat{t}_1$, we have

$$\frac{\alpha}{n} < q^n_{\lceil \hat{t} g_n^{-2} \rceil}(x, x_i) < \frac{K}{n}.$$

Choose $C, \beta, \hat{t}_2, n_2$ by Theorem S2.5 so that for $n > n_2, \hat{t} > \hat{t}_2$, we have

$$|q^n_{\lceil \hat{t} g_n^{-2} \rceil}(x, x_i) - \pi_{X_n}(x)| < C \exp(-\beta \hat{t}) \pi_{X_n}(x).$$

Thus, for $n > \max\{n_1, n_2\}$ and $\hat{t} > \max\{\hat{t}_1, \hat{t}_2\}$, we have

$$|q^n_{\lceil \hat{t} g_n^{-2} \rceil}(x, x_i) - \pi_{X_n}(x)| < C \exp(-\beta \hat{t}) \pi_{X_n}(x) < \frac{CK}{n} \exp(-\beta \hat{t}). \tag{1}$$

Now, for any $\gamma > 0$, choose $n_0 > \max\{n_1, n_2\}$ and $\hat{t}_0 > \max\{\hat{t}_1, \hat{t}_2\}$ large enough and $\delta > 0$ so that

- $\frac{CK}{n_0} \exp(-\beta \hat{t}_0) < \gamma/3$;
- by eventual uniform equicontinuity of $n q^n_{\lceil \hat{t}_0 g_n^{-2} \rceil}(x, x_i)$, for $n > n_0$, if $|x - y| < \delta$, then

$$|n q^n_{\lceil \hat{t}_0 g_n^{-2} \rceil}(x, x_i) - n q^n_{\lceil \hat{t}_0 g_n^{-2} \rceil}(y, x_i)| < \gamma/3.$$

Now, for $n > n_0$ and $|x - y| < \delta$, we find that

$$\begin{aligned}
|n\pi_{X^n}(x) - n\pi_{X^n}(y)| &\leq |n q^n_{\lceil \hat{t}_0 g_n^{-2} \rceil}(x, x_i) - n q^n_{\lceil \hat{t}_0 g_n^{-2} \rceil}(y, x_i)| + |n q^n_{\lceil \hat{t}_0 g_n^{-2} \rceil}(x, x_i) - n\pi_{X^n}(x)| \\
&\quad + |n q^n_{\lceil \hat{t}_0 g_n^{-2} \rceil}(y, x_i) - n\pi_{X^n}(y)| \\
&< \gamma/3 + \frac{2CK}{n} \exp(-\beta \hat{t}_0) \\
&< \gamma,
\end{aligned}$$

where we apply (1). This implies that $n\pi_{X^n}(x)$ is eventually uniformly equicontinuous, as needed. □

## 2.3  Expected hitting times degenerate to the stationary distribution

For any $x_j$, let $\pi'_{X^n}$ be the stationary distribution of the simple random walk on the graph $G'_n$ formed from $G_n$ by removing $x_j$ and all edges incident to it.

**Lemma S2.7.** *Assuming ($\star$), the rescaled stationary density $n\pi'_{X^n}(x)$ is a.s. eventually uniformly equicontinuous and satisfies*

$$\lim_{n \to \infty} n\pi'_{X^n}(x) = \hat{\pi}(x).$$

*Proof.* Let $q'_t(x, x_i)$ be the marginal distribution of the simple random walk on the modified graph $G'_n$. Because $G'_n$ is also a spatial graph, by (Hashimoto et al., 2015, Theorem 3.4), the time-rescaled simple random walks on $G'_n$ and $G_n$ converge to the same continuous-time Itô process, and we have under ($\star$) that

$$\lim_{n \to \infty} n\pi'_{X^n}(x_i) = \hat{\pi}(x_i),$$

where the convergence is uniform in $x_i$. □

**Lemma S2.8.** *There exist $t_0 > 0$, $n_0 > 0$, and $C, \beta > 0$ so that for all $\hat{t} > t_0$ and $n > n_0$ and any integer $t > \hat{t} g_n^{-2}$, we have*

$$\left| n \mathbb{P}\left( T_{x_j,n}^{x_i} = t \mid T_{x_j,n}^{x_i} \geq t \right) - n \sum_{x \in \mathsf{NB}_n^{in}(x_j)} \frac{\pi'_{X^n}(x)}{|\mathsf{NB}_n(x)|} \right| < C \exp(-\beta t g_n^2).$$

*Proof.* By Theorem S2.5, we may choose $\hat{t}_0 > 0$, $n_0 > 0$, $C_1 > 0$, and $\beta > 0$ so that for $\hat{t} > \hat{t}_0$ and $n > n_1$, we have

$$|q'_{\lceil \hat{t} g_n^{-2} \rceil - 1}(x, x_i) - \pi'_{X^n}(x)| < C_1 \exp(-\beta \hat{t}) \pi'_{X^n}(x_j).$$

We claim that the desired result will hold for $\hat{t}_0$ and this $n_0$.

By definition, we have

$$\mathbb{P}\left( T_{x_j,n}^{x_i} = t \mid T_{x_j,n}^{x_i} \geq t \right) = \sum_{x \in \mathsf{NB}_n^{in}(x_j)} \frac{q'_{t-1}(x, x_i)}{|\mathsf{NB}_n(x)|}$$

from which we conclude that for $t > \hat{t}_0 g_n^{-2}$ we have

$$\begin{aligned}
\left| \mathbb{P}\left( T_{x_j,n}^{x_i} = t \mid T_{x_j,n}^{x_i} \geq t \right) - \sum_{x \in \mathsf{NB}_n^{in}(x_j)} \frac{\pi'_{X^n}(x)}{|\mathsf{NB}_n(x)|} \right| &\leq \left| \sum_{x \in \mathsf{NB}_n^{in}(x_j)} \frac{q'_{t-1}(x, x_i) - \pi'_{X^n}(x)}{|\mathsf{NB}_n(x)|} \right| \\
&< C_1 \exp(-\beta t g_n^2) \sum_{x \in \mathsf{NB}_n^{in}(x_j)} \frac{\pi'_{X^n}(x)}{|\mathsf{NB}_n(x)|} \\
&< C_1 \exp(-\beta t g_n^2) \frac{|\mathsf{NB}_n^{in}(x_j)|}{\min_x |\mathsf{NB}_n(x)|} \max_x \pi'_{X^n}(x).
\end{aligned}$$

We now show there exists $C_2 > 0$ such that $\frac{|\mathsf{NB}_n^{in}(x_j)|}{\min_x |\mathsf{NB}_n(x)|} < C_2$ almost surely due to the construction of $g_n$ and $\bar{\varepsilon}$. By the out-degree estimate of an isotropic graph (Hashimoto et al., 2015, Theorem S3.2)[1], we have

$$\frac{|\mathsf{NB}_n(x)|}{|\mathcal{X}_n \cap B(x, \varepsilon_n(x))|} \to C(h) p(x)$$

for some constant $C(h)$ independent of $x$ and $n$. Further, since the number of points in $|\mathcal{X}_n \cap B(x, \varepsilon_n(x))| \sim \mathrm{Pois}(\varepsilon_n(x)^d V_d p(x))$, we obtain for a constant $0 < C_x < \infty$ dependent on $p, V_d$ and $C(h)$ that

$$\frac{|\mathsf{NB}_n(x)|}{\varepsilon_n(x)^d n} \to C_x.$$

For the denominator $\min_x |\mathsf{NB}_n(x)|$, the above limit immediately implies that $\min_x |\mathsf{NB}_n(x)| \varepsilon_n^{-d} n^{-1} \to \min_x C_x > 0$ by the lower bounds on $p(x)$ and $\varepsilon_n(x)$.

For the numerator, note that by construction of $\bar{\varepsilon}$, for any $\delta > 0$ there exists a $n$ such that $\varepsilon_n(x)^{-d} < (1 + \delta) \max_x \bar{\varepsilon}(x)^{-d} g_n^{-d}$ almost surely. By the expectation in (Hashimoto et al., 2015, Theorem S3.2), the out-neighborhood of a graph constructed with uniform scale $\max_x \bar{\varepsilon}(x) g_n$ asymptotically dominate the in-neighborhood of the original graph. Therefore,

$$\max_x |\mathsf{NB}_n(x)^{in}| \bar{\varepsilon}(x)^{-d} g_n^{-d} n^{-1} < \max_x C_x (1 + \delta) < \infty.$$

Combining the two bounds gives that

$$\frac{|\mathsf{NB}_n^{in}(x_j)|}{\min_x |\mathsf{NB}_n(x)|} < (1 + \delta) \frac{\max_x C_x \bar{\varepsilon}(x)^d}{\min_x C_x \bar{\varepsilon}(x)^d}.$$

The ratio $\frac{\max_x C_x}{\min_x C_x}$ is bounded by definition of $p(x)$, and therefore there exists $C_2 > 0$ such that $\frac{|\mathsf{NB}_n^{in}(x_j)|}{\min_x |\mathsf{NB}_n(x)|} < C_2$ almost surely. Finally, by Lemma S2.7, there exists $C_3 > 0$ such that $\pi'_{X^n}(x) \leq C_3/n$ for large enough $n$. The original statement follows by setting $C = C_1 C_2 C_3$. $\square$

---

[1] Note that there is a typographical error in Hashimoto et al. (2015) adding an additional factor of $\varepsilon_n(x)^{-d}$.

**Lemma S2.9.** *We have the limit*

$$\lim_{n \to \infty} \sum_{x \in \mathsf{NB}_n^{\mathrm{in}}(x_j)} \frac{1}{|\mathsf{NB}_n(x)|} = 1.$$

*Proof.* We will proceed through three estimates.

**Estimating $\overline{\varepsilon}(x)$ for $x \in \mathsf{NB}_n^{\mathrm{in}}(x_j)$:** For $\sigma > 0$, define $\gamma = \sigma \min_x \overline{\varepsilon}(x) > 0$. We may choose $\delta > 0$ so that if $|x - y| < \delta$, then $|\overline{\varepsilon}(x) - \overline{\varepsilon}(y)| < \gamma$. Choose $n_0$ so that if $n > n_0$ then $g_n \max_x \overline{\varepsilon}(x) < \delta/2$. For $n > n_0$, we find that for $x \in \mathsf{NB}_n^{\mathrm{in}}(x_j)$, we have

$$|x - x_j| \leq \varepsilon_n(x) \leq g_n \max_x \overline{\varepsilon}(x) < \delta$$

and therefore that

$$|\overline{\varepsilon}(x) - \overline{\varepsilon}(x_j)| < \sigma \min_x \overline{\varepsilon}(x) \leq \sigma \overline{\varepsilon}(x_j).$$

This implies that for $n > n_0$ we have

$$(1 - \sigma)\overline{\varepsilon}(x_j) < \overline{\varepsilon}(x) < (1 + \sigma)\overline{\varepsilon}(x_j). \tag{2}$$

**Estimating $|\mathsf{NB}_n(x)|$ for $x \in \mathsf{NB}_n^{\mathrm{in}}(x_j)$:** By (Hashimoto et al., 2015, Theorem S3.2), we have

$$\frac{|\mathsf{NB}_n(x)|}{|\mathcal{X}_n \cap B(x, \varepsilon_n(x))|} \to C(h)p(x)$$

for some constant $C(h)$ independent of $x$ and $n$.[2] For any $\tau > 0$, we may therefore find some $n_1$ so that for $n > n_1$ we have

$$(1 - \tau)C(h)p(x)|\mathcal{X}_n \cap B(x, \varepsilon_n(x))| < |\mathsf{NB}_n(x)| < (1 + \tau)C(h)p(x)|\mathcal{X}_n \cap B(x, \varepsilon_n(x))|.$$

On the other hand, by (2), for $x \in \mathsf{NB}_n^{\mathrm{in}}(x_j)$ and any $\sigma > 0$ there is some $n_0$ so that for $n > n_0$ we have

$$|\mathcal{X}_n \cap B(x, (1 - \sigma)\varepsilon_n(x_j))| < |\mathcal{X}_n \cap B(x, \varepsilon_n(x))| < |\mathcal{X}_n \cap B(x, (1 + \sigma)\varepsilon_n(x_j))|. \tag{3}$$

**Estimating $|\mathsf{NB}_n^{\mathrm{in}}(x_j)|$:** By (2) and an analogue of the proof of (Hashimoto et al., 2015, Theorem S3.2), we have for $x \in \mathsf{NB}_n^{\mathrm{in}}(x_j)$ that for any $\rho > 0$, there is $n_2 > 0$ so that if $n > n_2$ then

$$(1 - \rho)C(h)p(x)|\mathcal{X}_n \cap B(x, (1 - \sigma)\varepsilon_n(x_j))| < |\mathsf{NB}_n^{\mathrm{in}}(x_j)| < (1 + \rho)C(h)p(x)|\mathcal{X}_n \cap B(x, (1 + \sigma)\varepsilon_n(x_j))|. \tag{4}$$

**Completing the proof:** The conclusion follows by taking $\tau, \sigma, \rho \to 0$, choosing $n$ large, and combining (3) and (4). $\qquad\square$

**Lemma S2.10.** *The quantity* $\theta_n(x_j) = \sum_{x \in \mathsf{NB}_n^{\mathrm{in}}(x_j)} \frac{\pi'_{\mathcal{X}^n}(x)}{|\mathsf{NB}_n(x)|}$ *satisfies*

$$\lim_{n \to \infty} n\theta_n(x_j) = \widehat{\pi}(x_j).$$

*Proof.* Fix a sequence of points $y_1, y_2, \ldots$ in $\mathcal{X}$ with $y_k \in G'_k$ so that $\lim_{k \to \infty} y_k = x_j$. Fix any $\delta > 0$. By Lemma S2.7, we may find some $n_0$ so that for $n > n_0$, for each $x \in \mathsf{NB}_n^{\mathrm{in}}(x_j)$ we have

$$|\pi'_{\mathcal{X}^n}(x) - \pi'_{\mathcal{X}^n}(y_n)| < \delta/2.$$

This implies that for $n > n_0$ we have

$$\left| n\theta_n(x_j) - n\pi'_{\mathcal{X}^n}(y_n) \sum_{x \in \mathsf{NB}_n^{\mathrm{in}}(x_j)} \frac{1}{|\mathsf{NB}_n(x)|} \right| < \frac{\delta}{2} \sum_{x \in \mathsf{NB}_n^{\mathrm{in}}(x_j)} \frac{1}{|\mathsf{NB}_n(x)|}.$$

The result then follows by Lemma S2.9 and Lemma S2.7. $\qquad\square$

---

[2]Note that there is a typographical error in Hashimoto et al. (2015) adding an additional factor of $\varepsilon_n(x)^{-d}$.

**Theorem S2.11.** *For any $x_i$ and $x_j$, we have*

$$\frac{\mathbb{E}[T_{x_j,n}^{x_i}]}{n} \to \frac{1}{\widehat{\pi}(x_j)},$$

*where the convergence is a.s. in the draw of $\mathcal{X}$.*

*Proof.* By definition, we have

$$\mathbb{E}[T_{x_j,n}^{x_i} \mid T_{x_j,n}^{x_i} > \widehat{t}g_n^{-2}] \geq \mathbb{E}[T_{x_j,n}^{x_i}] \geq \mathbb{P}(T_{x_j,n}^{x_i} > \widehat{t}g_n^{-2})\mathbb{E}[T_{x_j,n}^{x_i} \mid T_{x_j,n}^{x_i} > \widehat{t}g_n^{-2}]. \tag{5}$$

By Lemma 3.1, for any $\delta > 0$ and $\widehat{t}_0 > 0$, there is some $n_1$ so that for $n > n_1$ and $\widehat{t} > \widehat{t}_0$ we have $\mathbb{P}(T_{x_j,n}^{x_i} > \widehat{t}g_n^{-2}) > (1 - \delta)$. Define now $p_t = \mathbb{P}(T_{x_j,n}^{x_i} = t \mid T_{x_j,n}^{x_i} \geq t)$; by definition we have

$$\mathbb{E}[T_{x_j,n}^{x_i} \mid T_{x_j,n}^{x_i} > \widehat{t}g_n^{-2}] = \sum_{t=\lceil\widehat{t}g_n^{-2}\rceil}^{\infty} t p_t \prod_{r=\lceil\widehat{t}g_n^{-2}\rceil}^{t-1} (1 - p_r).$$

By Lemma S2.8, we have for some $n_2$ that for $n > n_2$ and $t > \widehat{t}_0 g_n^{-2}$ that

$$|p_t - \theta_n(x_j)| < \frac{C \exp(-\beta t g_n^2)}{n}$$

so in particular for $\delta = \frac{1}{2}\min_{x \in D}\widehat{\pi}(x)$ and $\tau = 2\max_{x \in D}\widehat{\pi}(x)$, we have for some $n_3$ that for $n > n_3$ we have

$$\delta < np_t < \tau \text{ and } \delta < n\theta_n(x_j) < \tau.$$

For $n_4$ large enough that $1 - \tau/n_4 > \delta/n_4$, for $n > n_4$ we have

$$\left| p_t \prod_{r=\lceil\widehat{t}g_n^{-2}\rceil}^{t-1} (1 - p_r) - \theta_n(x_j)(1 - \theta_n(x_j))^{t-\lceil\widehat{t}g_n^{-2}\rceil} \right| < \sum_{r=\lceil\widehat{t}g_n^{-2}\rceil}^{t-1} \frac{C\exp(-\beta r g_n^2)}{n}(1 - \tau/n)^{t-\lceil\widehat{t}g_n^{-2}\rceil - 1}$$

$$< \frac{C}{n}\frac{e^{-\beta\widehat{t}}}{1 - e^{-\beta g_n^2}}(1 - \tau/n)^{t-\lceil\widehat{t}g_n^{-2}\rceil - 1}.$$

This implies that

$$\frac{1}{n}\left| \mathbb{E}[T_{x_j,n}^{x_i} \mid T_{x_j,n}^{x_i} > \widehat{t}g_n^{-2}] - \sum_{t=\lceil\widehat{t}g_n^{-2}\rceil}^{\infty} t\theta_n(x_j)(1 - \theta_n(x_j))^{t-\lceil\widehat{t}g_n^{-2}\rceil} \right| < \sum_{t=\lceil\widehat{t}g_n^{-2}\rceil}^{\infty} \frac{C}{n^2}\frac{e^{-\beta\widehat{t}}}{1 - e^{-\beta g_n^2}}(1 - \tau/n)^{t-\lceil\widehat{t}g_n^{-2}\rceil - 1}$$

$$< \frac{C}{\tau(n-\tau)}\frac{e^{-\beta\widehat{t}}}{1 - e^{-\beta g_n^2}},$$

where we note that for $n > 2\tau$, we have

$$\frac{C}{\tau(n-\tau)}\frac{e^{-\beta\widehat{t}}}{1 - e^{-\beta g_n^2}} < \frac{2Ce^{-\beta\widehat{t}}}{\tau}n^{-1}(g_n^{-2} + \frac{1}{2} + \frac{1}{12}g_n^2).$$

Because $\lim_{n\to\infty} n^{-1}(g_n^{-2} + \frac{1}{2} + \frac{1}{12}g_n^2) = 0$, considering $n > \max\{n_1, n_2, n_3, n_4\}$, we conclude that

$$\lim_{n\to\infty} \frac{1}{n}\mathbb{E}[T_{x_j,n}^{x_i} \mid T_{x_j,n}^{x_i} > \widehat{t}g_n^{-2}] = \lim_{n\to\infty} \frac{1}{n}\sum_{t=\lceil\widehat{t}g_n^{-2}\rceil}^{\infty} t\theta_n(x_j)(1 - \theta_n(x_j))^{t-\lceil\widehat{t}g_n^{-2}\rceil}$$

$$= \lim_{n\to\infty} \frac{1}{n}\frac{1 - \theta_n(x_j) + \theta_n(x_j)\lceil\widehat{t}g_n^{-2}\rceil}{\theta_n(x_j)}$$

$$= \lim_{n\to\infty} \frac{1}{n\theta_n(x_j)} = \frac{1}{\widehat{\pi}(x_j)},$$

where the last equality follows from Lemma S2.10. Now by (5), we conclude that

$$\lim_{n\to\infty} \frac{1}{n}\mathbb{E}[T_{x_j,n}^{x_i}] = \frac{1}{\widehat{\pi}(x_j)}. \qquad \square$$

## 2.4 The case of one dimension

The Laplacian-based bounds in (von Luxburg et al., 2014) suggest that the hitting time should diverge even when the dimension of the underlying geometric graph is 1. This is a very surprising result, since the continuous random walk in one dimension converges to a non-trivial limit. We provide another explanation of this result in our framework.

Intuitively, this happens since we are concerned with the hitting time to a single point, and the discrete random walk may jump over the point, while the continuous walk cannot. To demonstrate this, we show that considering the hitting time to a sufficiently large out-neighborhood of a vertex instead of the vertex itself fixes this problem.

Pick $x_i, x_j \in G_n$, and let $X_t^n$ be the simple random walk on $G_n$. Suppose without loss of generality that $x_i < x_j$ and define

$$\gamma = \inf_n \min_{x_i \in \mathcal{X}_n} x_i$$

to be the left boundary of $D$. Pick a sequence of sets of vertices $S_n \subset \mathcal{X}_n$ so that every element in $S_n$ is reachable from $x_j$ in $o(g_n^{-1})$ steps and the removal of $S_n$ from $G_n$ disconnects $G_n$. Let $T_{S_n}^{x_i}$ be the hitting time to any point in $S_n$. We will use the Feynman-Kac theorem for functionals of hitting time.

**Theorem S2.12** ((Øksendal, 2003, Exercise 9.12) Feynman-Kac). *Let $Z_t$ be an Itô process in $\mathbb{R}^d$ defined by*

$$dZ_t = \mu(Z_t)dt + \sigma(Z_t)dB_t.$$

*For a function $f(x)$ and $T_E^x$ the hitting time to a domain $E \subset D$, the function*

$$u(x) = \mathbb{E}\left[\int_0^{T_E^x} f(Z_s)ds\right]$$

*is the solution to the boundary value problem*

$$\frac{1}{2}Tr[\sigma^T Hu\sigma] + \mu(x) \cdot \nabla u + f(x) = 0$$

*with boundary condition $u|_{\partial E} = 1$.*

**Theorem S2.13.** *Such a sequence of vertex sets $S_n$ always exists and the expected hitting time $\mathbb{E}[T_{S_n,n}^{x_i}]$ converges to a non-degenerate continuum limit defined by*

$$\mathbb{E}[T_{S_n,n}^{x_i}g_n^2] \to \int_{x_i}^{x_j} \frac{1}{p(y)^2} \int_{\gamma}^y \frac{2p(z)^2}{\bar{\varepsilon}(z)^2} dz dy.$$

*Proof.* First we prove a sequence $S_n$ exists. Take the set of points $\widehat{S}_n = \{x_k : |x_k - x_j| < c_n\}$ for a sequence $c_n$ with $c_n \to 0$ and $c_n g_n \to \infty$. Let $s$ be the maximum shortest path distance to any element in $\widehat{S}_n$. Then we have $s = o(g_n^{-1})$ since $c_n \to 0$ and the length of the shortest path between any two points scales as $\Theta(g_n^{-1})$. Therefore the set $S_n$ defined by all points whose shortest path distance to $x_j$ is at most $s$ fulfills the requirements.

Let $\widehat{T}_{ij}$ be the hitting time to $x_j$ of $Y_{\widehat{t}}$ started at $x_i$. Note that it is not infinite because we have $d = 1$. By Corollary 4.4 and the fact that $sg_n^{-1} \to 0$, we have

$$T_{S_n,n}^{x_i}g_n^2 \xrightarrow{d} \widehat{T}_{ij}.$$

Finally, by Theorem S2.12 with $f(x) \equiv 1$, the expected hitting time $u(x)$ to $x_j$ under the continuous process $Y_{\widehat{t}}$ started at $x$ is the solution to the boundary value problem

$$\frac{1}{2}\bar{\varepsilon}(x)^2 u''(x) + \frac{p'(x)}{p(x)}\bar{\varepsilon}(x)^2 u'(x) + 1 = 0$$

We may rewrite this as

$$p(x)^2 u''(x) + 2p(x)p'(x)u'(x) = -\frac{2p(x)^2}{\bar{\varepsilon}(x)^2},$$

after which integration of both sides and application of $u'(\gamma) = 0$ implies that

$$p(x)^2 u'(x) = -\int_\gamma^x \frac{2p(z)^2}{\overline{\varepsilon}(z)^2} dz.$$

Another integration and application of $u(x_j) = 0$ implies that

$$u(x) = -\int_{x_j}^x \frac{1}{p(y)^2} \int_\gamma^y \frac{2p(z)^2}{\overline{\varepsilon}(z)^2} dz dy.$$

Setting $x = x_i$ then implies that

$$\lim_{n\to\infty} \mathbb{E}[T_{S_n,n}^{x_i} g_n^2] = \mathbb{E}[\widehat{T}_{ij}] = \int_{x_i}^{x_j} \frac{1}{p(y)^2} \int_\gamma^y \frac{2p(z)^2}{\overline{\varepsilon}(z)^2} dz dy. \qquad \square$$

For cases where the kernel function takes values in $\{0, 1\}$, such as the $k$-nearest neighbor graph, the following corollary is useful.

**Corollary S2.14.** *Suppose that $G_n$ is constructed by the kernel $h(x) = 1_{[0,1]}$. Then the expected hitting time of $X_t^n$ started at $x_i$ to the out-neighbors of $x_j$ converges to the limit of Theorem S2.13*

*Proof.* From the fact that the out-neighborhood of $x_j$ satisfies the conditions for $S_n$ in Theorem S2.13. $\square$

Although this metric is nontrivial in the sense that it retains some information about the latent space metric, it is still highly distorted. We examine this phenomenon in the case of $\overline{\varepsilon}(x) = 1$ and $p(x) = 1$ in the following Corollary.

**Corollary S2.15.** *If $\overline{\varepsilon}(x) = 1$ and $p(x) = 1$ in Corollary S2.14, for any $x_i$ and $x_j$ the rescaled expectation of the hitting time $T_{\mathsf{NB}_n(x_j),n}^{x_i}$ of $X_t^n$ started at $x_i$ to the out-neighborhood of $x_j$ has the limit*

$$\mathbb{E}[T_{\mathsf{NB}_n(x_j),n}^{x_i} g_n^2] \to |x_j - x_i| \cdot |x_j + x_i - 2\gamma|.$$

*Proof.* This follows by applying Theorem S2.13 with our $\overline{\varepsilon}(x)$ and $p(x)$. $\square$

*Remark 2.* Note that the boundary condition in Corollary 2.15 induces a large non-uniform multiplicative error. Because of this, the expected hitting time is not consistent even in the ideal situation of a one-dimensional latent space with random walk converging to Brownian motion. Compare this result with Theorem 4.5, which shows a much stronger consistency property.

# 3 Computing the LTHT

Algorithmically, computing the LTHT can be done in two major ways: matrix inversion, or sampling. For the results in the paper, we use the direct sampling method of drawing a simple random walk and calculating the exponentially discounted hitting time. This same computation can be performed using a truncated power method (Yazdani, 2013, Algorithm 1).

Alternative approaches for computing the LTHT involve the following matrix inversion method. Let $P$ be the transition matrix for some random walk. Then the LTHT $\mathbb{E}[\exp(-\beta T_{x_j,n}^{x_i})]$ is given by

$$\mathbb{E}[\exp(-\beta T_{x_j,n}^{x_i})] = (I - W \exp(-\beta))_{ji}^{-1}.$$

Note that this expression is a close discrete analog of Feynman-Kac (Theorem 2.1). This relationship was used in prior work (Smith et al., 2014, Eq. 22) to calculate the LTHT in a different setting and formulation. Correctness of this expression can be seen via the series expansion which was computed as a normalizer for randomized shortest paths (Françoisse et al., 2013, Algorithm 2). This method has been used to calculate the LTHT in in prior work (Kivimäki et al., 2014).

# 4 Reweighting the random walk

Recall that $A_{ij}^n$ was the adjacency matrix of $G_n$. In Corollary S4.2, we give a complete proof of Theorem 4.1 from the maintext.

## 4.1 General construction and application to Brownian motion

Let $a_n(x)$ and $b_n(x)$ be scalar functions on $\mathcal{X}_n$ with possibly stochastic dependence on $\mathcal{X}_n$ so that

$$\lim_{n \to \infty} a_n(x) = \bar{a}(x) \qquad \text{and} \qquad \lim_{n \to \infty} b_n(x) = \bar{b}(x)$$

uniformly in $x$ for some deterministic $\bar{a}(x)$ and $\bar{b}(x)$.

**Theorem S4.1.** *If $a_n(x)$ is a.s. eventually equicontinuous, $\bar{a}(x)$ is smooth with bounded gradient, and $\bar{b}(x)$ is continuous and bounded in $(0, 1]$, the weighted random walk $Z_t$ defined by the transition matrix*

$$\mathbb{P}(Z_{t+1} = x_j \mid Z_t = x_i) = \begin{cases} A_{i,j}^n \frac{a_n(x_j)}{\sum_{x_k \in \mathsf{NB}_n(x_i)} a_n(x_k)} b_n(x_i) & i \neq j \\ 1 - b_n(x_i) & i = j \end{cases}$$

*converges to the Itô process with drift $\nabla \log(p(x)\bar{a}(x))/3$ and diffusion $\bar{\varepsilon}(x)^2\bar{b}(x)/3$.*

*Proof.* To show convergence to an Itô process, it suffices to check the Stroock-Varadhan criterion (Stroock & Varadhan, 1971). Since the boundary for both the original and modified walk are the same, we only need check that

$$\mathbb{E}[Z_{t+1} - x_i \mid Z_t = x_i] \xrightarrow{p} \frac{1}{3} \frac{\nabla[p(x_i)\bar{a}(x_i)]}{p(x_i)\bar{a}(x_i)} \bar{\varepsilon}(x_i)^2 \bar{b}(x_i), \text{ and}$$

$$\mathbb{E}[(Z_{t+1} - x_i)^2 \mid Z_t = x_i] \xrightarrow{p} \frac{1}{3} \bar{\varepsilon}(x_i)^2 \bar{b}(x_i).$$

For this, by definition we have that

$$\mathbb{E}[Z_{t+1} - x_i \mid Z_t = x_i] = \mathbb{P}(Z_{t+1} \neq x_i) \frac{1}{\sum_{x_k \in \mathsf{NB}_n(x_i)} a_n(x_k)} \sum_{x_k \in \mathsf{NB}_n(x_i)} (x_k - x_i) a_n(x_k), \text{ and}$$

$$\mathbb{E}[(Z_{t+1} - x_i)^2 \mid Z_t = x_i] = \mathbb{P}(Z_{t+1} \neq x_i) \frac{1}{\sum_{x_k \in \mathsf{NB}_n(x_i)} a_n(x_k)} \sum_{x_k \in \mathsf{NB}_n(x_i)} (x_k - x_i)^2 a_n(x_k),$$

from which the desired estimates follow by using $\mathbb{P}(Z_{t+1} \neq x_i) = b_n(x_i)$ and the values and concentration of conditional moments $\mathbb{E}[f(Z_t - Z_{t-1}) \mid Z_{t-1}, Z_t \neq Z_{t-1}]$ given by applying Lemma S4.3 and Lemma S4.4 to $f(x) = x$ and $f(x) = x^2$. □

**Corollary S4.2.** *Let $\hat{p}$ and $\hat{\varepsilon}$ be consistent estimators of the density and local scale and $A$ be the adjacency matrix. Then the random walk $\hat{X}_t^n$ defined by the following transition*

$$\mathbb{P}(\hat{X}_{t+1}^n = x_j \mid \hat{X}_t^n = x_i) = \begin{cases} \frac{A_{i,j}\hat{p}(x_j)^{-1}}{\sum_k A_{i,k}\hat{p}(x_k)^{-1}} \hat{\varepsilon}(x_i)^{-2} & i \neq j \\ 1 - \hat{\varepsilon}(x_i)^{-2} & i = j \end{cases}$$

*converges to a Brownian motion.*

*Proof.* Set $a_n(x) = \hat{p}(x)^{-1}$ and $b_n(x) = \hat{\varepsilon}(x)^{-2}$ as estimated by (Hashimoto et al., 2015) so that $\lim_{n \to \infty} a_n(x) = p(x)^{-1}$ and $\lim_{n \to \infty} b_n(x) = \bar{\varepsilon}(x)^{-2}$. These satisfy the conditions of Theorem S4.1 and yield limiting drift and diffusion coefficients for Brownian motion. □

## 4.2   Technical moment estimates

In this subsection, we give the moment estimates necessary in the proof of Theorem S4.1. We first derive the expected values of each moment quantity averaged over draws of $\mathcal{X}_n$.

**Lemma S4.3** (Expected values of reweighting). *Let $x = X_t^n$ and $y = X_{t+1}^n$. Then the conditional expectation after weighting by $a_n(x)$ converges to the weighted draw over $p(x)a_n(x)$; that is, we have a.s. that*

$$\lim_{n \to \infty} \left| \frac{1}{h_n} |\mathsf{NB}_n(x)| \, \mathbb{E}\left[ \frac{a_n(y)}{\sum_{z \in \mathsf{NB}_n(x)} a_n(z)} f(y-x) \mid y \neq x \right] \right.$$
$$\left. - \frac{1}{h_n} \int_{y \in B(x, \varepsilon_n(x))} f(y-x) \frac{p(y)\overline{a}(y)}{\int_{z \in B(x, \varepsilon_n(x))} p(z)\overline{a}(z)dz} dy \right| = 0.$$

*Proof.* By the continuity of $p$ and a.s. eventual equicontinuity of $a_n(y)$, we have $\sup_{y \in B(x, \varepsilon_n(x))} |a_n(y)p(y) - a_n(x)p(x)| \to 0$ and $\sup_{y \in B(x, \varepsilon_n(x))} |p(y) - p(x)| \to 0$. These together imply

$$\frac{\int_{y \in B(x, \varepsilon_n(x))} a_n(y)p(y)dy}{\int_{y \in B(x, \varepsilon_n(x))} p(y)dy} \xrightarrow{a.s.} \overline{a}(x). \tag{6}$$

Because $a_n(x) \to \overline{a}(x)$ uniformly in $x$, for any $\delta > 0$, we may choose $n_0$ so that for $n > n_0$, we have $|a_n(x) - \overline{a}(x)| < \delta/2$ and $\varepsilon_n(x)$ is small enough so that if $|y - x| < \varepsilon_n(x)$, then $|\overline{a}(y) - \overline{a}(x)| < \delta/2$. For $n > n_0$, we then have

$$\sup_{z \in \mathsf{NB}_n(x)} |a_n(z) - \overline{a}(x)| \leq \sup_{z \in \mathsf{NB}_n(x)} |a_n(z) - \overline{a}(z)| + |\overline{a}(z) - \overline{a}(x)| < \delta.$$

This shows that $\sup_{z \in \mathsf{NB}_n(x)} |a_n(z) - \overline{a}(x)| \to 0$ and therefore

$$\frac{\sum_{z \in \mathsf{NB}_n(x)} a_n(z)}{|\mathsf{NB}_n(x)|} \xrightarrow{a.s.} \overline{a}(x). \tag{7}$$

Applying (6) and (7), we find that

$$\lim_{n \to \infty} \left| \frac{1}{h_n} |\mathsf{NB}_n(x)| \, \mathbb{E}\left[ \frac{a_n(y)}{\sum_{z \in \mathsf{NB}_n(x)} a_n(z)} f(y-x) \mid x \neq y \right] \right.$$
$$\left. - \frac{1}{h_n} \mathbb{E}\left[ a_n(y) f(y-x) \mid x \neq y \right] \frac{\int_{y \in B(x, \varepsilon_n(x))} p(y)dy}{\int_{y \in B(x, \varepsilon_n(x))} a_n(y)p(y)dy} \right| \to 0.$$

We apply the argument of (Hashimoto et al., 2015, Lemma 3.2) to this iterated expectation to obtain

$$\lim_{n \to \infty} \left| \frac{1}{h_n} \mathbb{E}\left[ a_n(y) f(y-x) \mid x \neq y \right] \frac{\int_{y \in B(x, \varepsilon_n(x))} p(y)dy}{\int_{y \in B(x, \varepsilon_n(x))} a_n(y)p(y)dy} \right.$$
$$\left. - \frac{1}{h_n} \int_{y \in B(x, \varepsilon_n(x))} f(y-x) \frac{p(y)\overline{a}(y)}{\int_{z \in B(x, \varepsilon_n(x))} p(z)\overline{a}(z)dz} dy \right| \to 0. \quad \square$$

Evaluating the integrals for $f(x) = x$ and $f(x) = x^2$ in Lemma S4.3 implies that the expected value of an increment of the reweighted walk across all draws of $\mathcal{X}_n$ limits to $\nabla \log[p(x)\overline{a}(x)]/3$ and the expected variance of the increment limits to $\overline{\varepsilon}(x)^2 \overline{b}(x)/3$. However, in order to apply the Stroock-Varadhan criteria we require that this hold with high probability over all draws of $\mathcal{X}_n$.

**Lemma S4.4** (Strong LLN for local moments). *For a function $f(x)$ such that $\sup_{x \in B(0,\varepsilon)} |f(x)| < \varepsilon$ for small $\varepsilon > 0$, we have a.s. that*

$$\lim_{n \to \infty} \left| \frac{1}{h_n} \sum_{y \in \mathsf{NB}_n(x)} \frac{a_n(y)}{\sum_{z \in \mathsf{NB}_n(x)} a_n(z)} f(y-x) - \frac{1}{h_n} \int_{y \in B(x, \varepsilon_n(x))} f(y-x) \frac{p(y)\overline{a}(y)}{\int_{z \in B(x, \varepsilon_n(x))} p(z)\overline{a}(z)dz} dy \right| = 0.$$

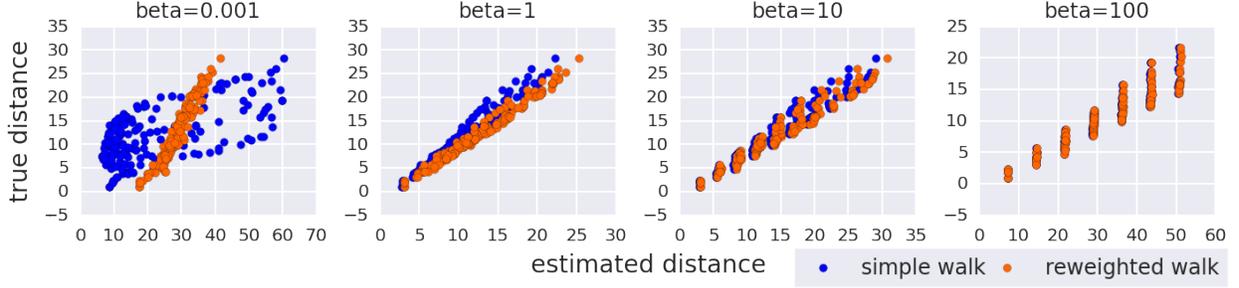

Figure 1: Distance estimates for various values of $\beta$ on re-weighted walks on a simulated dataset

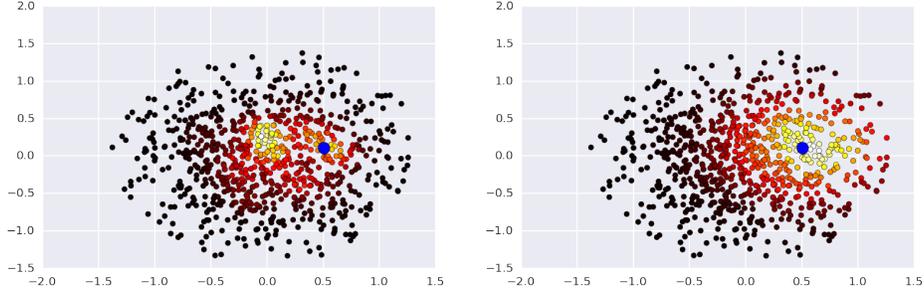

(a) Simple random walk is biased toward region with high density

(b) Re-weighted walk diffuses evenly on the true metric

Figure 2: Visualization of the marginal distribution $P_{ij}(t)$ of a random walk over a $k$-nn graph on a Gaussian restricted to a disk, starting at the blue initial point and run for 40 steps. The re-weighted walk diffuses evenly from the starting point, ignoring biases due to density $p$ and neighborhood size $\varepsilon$.

*Proof.* Define the quantity

$$\mu(x) = \frac{1}{h_n} \int_{y \in B(x, \varepsilon_n(x))} f(y - x) \frac{p(y)\overline{a}(y)}{\int_{z \in B(x, \varepsilon_n(x))} p(z)\overline{a}(z)dz} dy.$$

We wish to bound

$$p_n(t) = \mathbb{P}\left(\left| \frac{1}{h_n} \sum_{y \in \mathsf{NB}_n(x)} \frac{a_n(y)}{\sum_{z \in \mathsf{NB}_n(x)} a_n(z)} f(y - x) - \mu(x)\right| \ge t\right). \tag{8}$$

By a.s. eventual equicontinuity of $a_n(y)$, we have for some $c > 0$ and large enough $n$ that

$$\frac{a_n(y)}{\sum_{z \in \mathsf{NB}_n(x)} a_n(z)} \le c \frac{1}{|\mathsf{NB}_n(x)|}.$$

By the construction of $\varepsilon_n(x)$, if $|y - x| < \varepsilon_n(x)$, then $|f(y - x)| \le \varepsilon_n(x)$. Combining these two we apply Hoeffding's inequality to obtain that

$$p_n(t) \le 2 \exp\left(-\frac{2h_n^2 |\mathsf{NB}_n(x)|^2 c^2 t^2}{|\mathsf{NB}_n(x)|\varepsilon_n(x)^2}\right) = o(n^{-2t^2\omega(1)}), \tag{9}$$

where we use that $|\mathsf{NB}_n(x)| = \omega\left(n^{2/(d+2)} \log(n)^{d/(d+2)}\right)$. This completes the proof by Borel-Cantelli. $\qquad\square$

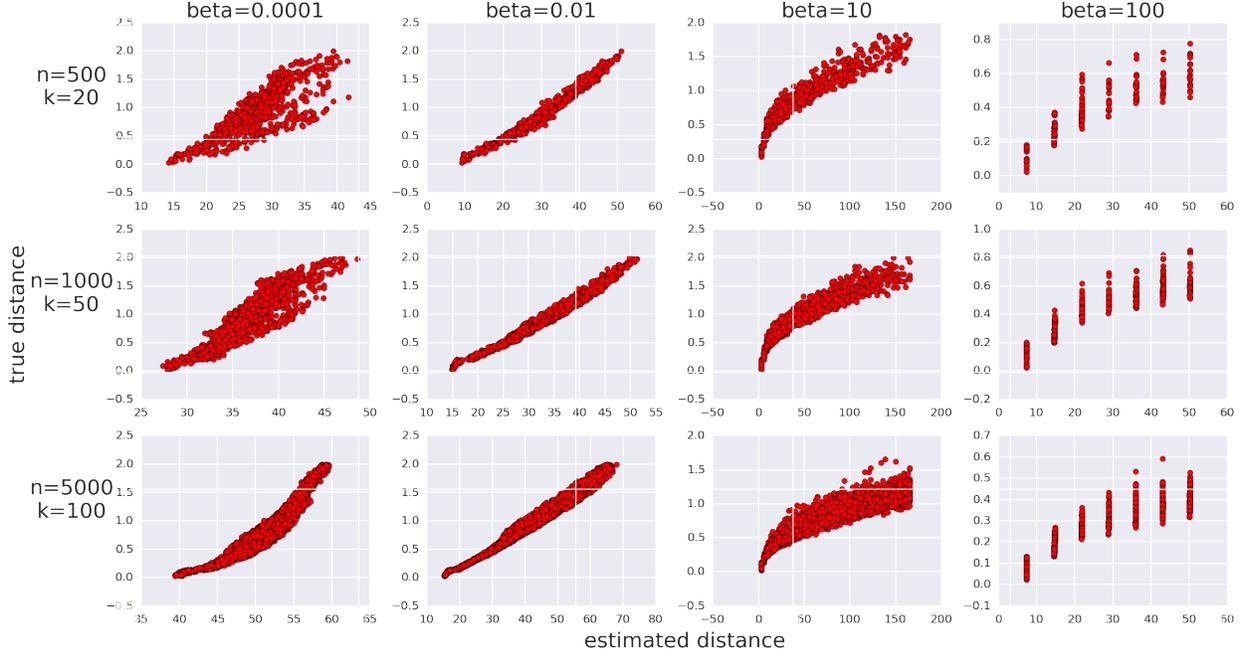

Figure 3: Distance estimates for various values of $\beta$ on re-weighted walks on a simulated dataset

# 5 Consistency at $\beta = \omega(\log(g_n^d n))$ via shortest paths

**Definition S5.1.** *Define the $f$-length of any path $\gamma \subset D$ as given in (Alamgir & von Luxburg, 2012) as*

$$D_{f,\gamma} = \int_\gamma f(\gamma(t))|\gamma'(t)|dt.$$

*Let the $f$-distance from $x$ to $y$ be the minimum path length between two points*

$$D_f(x,y) = \min_{\gamma \in C^1, \gamma(0)=x, \gamma(1)=y} D_{f,\gamma}.$$

**Theorem S5.2.** *Let $\beta = \omega(\log(g_n^d n))$, then for $f(x) = \overline{\varepsilon}(x)^{-1}$ we have*

$$-\log(\mathbb{E}[\exp(-\beta T^{x_i}_{x_j,n})])/\beta g_n \to D_f(x_i, x_j).$$

*Proof.* Define $H_{ij}(t)$ to be the probability of not hitting $x_j$ by step $t$, and $P_{ij}(t)$ to be the probability of going from $x_i$ to $x_j$ in exactly $t$ steps. The expected value is the series

$$-\log(\mathbb{E}[\exp(-\beta T^{x_i}_{x_j,n})])/\beta g_n = -\beta^{-1} g_n \log\left(\sum_{t=0}^\infty P_{ij}(t)H_{ij}(t)\exp(-\beta t)\right).$$

Now, let $D_{ij}$ be the length of the shortest path from $i$ to $j$. By definition $H_{ij}(D_{ij}) = 1$ and

$$-\log(\mathbb{E}[\exp(-\beta T^{x_i}_{x_j,n})])/\beta g_n = D_{ij}g_n - \log(P_{ij}(D_{ij}))\frac{g_n}{\beta} - \log\left(1 + \sum_{t=D_{ij}+1}^\infty \frac{P_{ij}(t)}{P_{ij}(D_{ij})}H_{ij}(t)\exp(-\beta(t-D_{ij}))\right)\frac{g_n}{\beta}.$$

This forms the upper bound

$$-\log(\mathbb{E}[\exp(-\beta T^{x_i}_{x_j,n})])/\beta g_n \le D_{ij}g_n - \log(P_{ij}(D_{ij}))\frac{g_n}{\beta}.$$

The probability $P_{ij}(D_{ij})$ of hitting $x_j$ in exactly $D_{ij}$ steps is lower bounded by $(g_n^d n)^{-D_{ij}}$ since by definition at least one path exists. This implies that $\log(P_{ij}(D_{ij})) = o(g_n^{-1} \log(g_n^d n))$ and therefore

$$D_{ij} g_n \leq -\log(\mathbb{E}[\exp(-\beta T_{x_j, n}^{x_i})])/\beta g_n \leq D_{ij} g_n - o(1),$$

where the lower bound follows because it is impossible to reach vertex $x_j$ in less than $D_{ij}$ steps. By (Alamgir & von Luxburg, 2012) for the $k$-nearest neighbor case and (Hashimoto et al., 2015) with Lemma S6.3 for the other cases of a spatial graph, $D_{ij} g_n$ converges to the $f$-distance defined by $\bar{\varepsilon}(x)^{-1}$, completing the proof. $\qquad\square$

# 6 Consistency of LTHT

In this section, we prove some results needed in the proof of Theorem 4.5.

## 6.1 LTHT of the Brownian motion

**Lemma S6.1.** *Let $W_t$ be a Brownian motion with $W_0 = x_i$. Let $\overline{T}_{B(x_j, s)}^{x_i}$ be the hitting time of $W_t$ to $B(x_j, s)$. For any $\alpha < 0$, if $\widehat{\beta} = s^\alpha$, as $s \to 0$ we have*

$$-\log(\mathbb{E}[\exp(-\widehat{\beta} \overline{T}_{B(x_j, s)}^{x_i})])/\sqrt{2\widehat{\beta}} \to |x_i - x_j|.$$

*Proof of Lemma 6.1.* Let $B_t = |W_t|$ be the order $\nu = d/2 - 1$ Bessel process. The LTHT of $B_t$ to hit $x_j \pm s$ is equivalent to the LTHT of $W_t$ to hit $B(x_j, s)$. Defining $w = |x_i - x_j|$, by (Borodin & Salminen, 2002, Eq 4.2.0.1), this is:

$$\mathbb{E}[\exp(-\widehat{\beta} \overline{T}_{B(x_j, s)}^{x_i})] = \frac{K(\nu, w\sqrt{2\widehat{\beta}}) w^{-\nu}}{K(\nu, s\sqrt{2\widehat{\beta}}) s^{-\nu}},$$

where $K(\nu, w)$ is a modified Bessel function of the second kind. Write $-\log(\mathbb{E}[\exp(-\widehat{\beta} \overline{T}_{B(x_j, s)}^{x_i})])/\sqrt{2\widehat{\beta}} = c_1 + c_2$ for

$$c_1 = -\log(K(\nu, w\sqrt{2\widehat{\beta}}) w^{-\nu})/\sqrt{2\widehat{\beta}}$$
$$c_2 = -\log(K(\nu, s\sqrt{2\widehat{\beta}}) s^{-\nu})/\sqrt{2\widehat{\beta}}.$$

Taylor expansion of $c_1$ at $\widehat{\beta}^{-1} = 0$ yields

$$c_1 = w - \frac{\log(\pi^2/(8\widehat{\beta})) + 4\log(w^{-1/2 - \nu})}{4\sqrt{2\widehat{\beta}}} + o\left(\frac{\nu^2}{w\widehat{\beta}}\right),$$

hence $c_1 \to w$. For $c_2$, note that $\nu \log(s)/\sqrt{2\widehat{\beta}} \to 0$ and for $s$ small,

$$K(\nu, s\sqrt{2\widehat{\beta}}) \sim \begin{cases} -\log(s\sqrt{2\widehat{\beta}}) & d = 2 \\ \frac{1}{2}\Gamma(s\sqrt{2\widehat{\beta}})(\frac{1}{2}s\sqrt{2\widehat{\beta}})^{-\nu} & d > 2 \end{cases}.$$

by (Abramowitz & Stegun, 1972, p375). Checking that $-\log(K(\nu, s\sqrt{2\widehat{\beta}}))/\sqrt{2\widehat{\beta}} \to 0$ and combining estimates gives $-\log(\mathbb{E}[\exp(-\widehat{\beta} \overline{T}_{B(x_j, s)}^{x_i})])/\sqrt{2\widehat{\beta}} = c_1 + c_2 \to w$. $\qquad\square$

## 6.2 Proof of Corollary 4.4

We prove here Corollary S6.2. We recall the setup. For points $x_i, x_j \in G_n$ and $s > 0$, $\widehat{T}^{x_i}_{B(x_j, s)}$ is the hitting time of the de-biased walk on $G_n$ from $x_i$ to $\mathsf{NB}^s_n(x_j)$. In the continuous setting, $\overline{T}^{x_i}_{B(x_j, s)}$ is the hitting time of Brownian motion with reflecting boundary conditions in $D$ from $x_i$ to $B(x_j, s)$. We would like to show the following.

**Corollary S6.2.** *For $s > 0$, we have $g_n^2 \widehat{T}^{x_i}_{B(x_j, s)} \xrightarrow{d} \overline{T}^{x_i}_{B(x_j, s)}$.*

Our proof consists of two steps. First, we show that hitting $\mathsf{NB}^s_n(x_j)$ is equivalent to hitting $B(x_j, s)$ with the discrete walk. Second, we use S4.2 to show convergence in distribution of this second hitting time. We require a few lemmas.

**Lemma S6.3.** *For any $\delta > 0$ and $s > 0$ so that $B(x_j, s + \delta) \subset D$, we have with high probability that*

$$\mathcal{X}_n \cap B(x_j, s - \delta) \subset \mathsf{NB}^s_n(x_j) \subset B(x_j, s + \delta).$$

*Proof.* Recall that $\mathsf{NB}^s_n(x_j)$ is defined as

$$\mathsf{NB}^s_n(x) := \{y \mid \text{there is a path } x \to y \text{ of } \widehat{\varepsilon}\text{-weight} \leq s\}.$$

The estimator $\widehat{\varepsilon}(x)$ is appropriately scaled such that $\widehat{\varepsilon}(x) \to \overline{\varepsilon}(x)$ uniformly and almost surely. Thus, we need to show that $\widehat{\varepsilon}$-weighted shortest path distance converges to true shortest path distance up to error $\Theta(g_n)$.

We first present the simpler case of a constant kernel $h(x) \equiv 1$ over $[0, 1]$; this includes the $k$-nearest neighbor and $\varepsilon$-ball cases. Let $D_{ij}$ be the minimum $\widehat{\varepsilon}$-weight of a path from $x_i$ to $x_j$. The proof of (Hashimoto et al., 2015, Theorem S4.5) shows that in this case

$$\left| |x_i - x_j| - D_{ij} g_n \right| \leq \varepsilon_n(x_j). \tag{10}$$

If $x_k \in \mathcal{X}_n \cap B(x_j, s - \delta)$, this implies that $D_{jk} g_n \to |x_k - x_j| \leq s - \delta$. Therefore $D_{jk} g_n \leq s$ with high probability and $x_k \in \mathsf{NB}^s_n(x_j)$. If $x_k \in \mathsf{NB}^s_n(x_j)$, this implies that $D_{ik} \leq s$. By Equation (6.2), we have $s \geq D_{ij} g_n \to |x_i - x_j|$. Therefore $x_k \in B(x_j, s + \delta)$ with high probability.

The proof for the case of generic $h(x)$ is closely analogous. The same proof as used for (Hashimoto et al., 2015, Theorem S4.5) shows that there exists some $k$ such that $|x_k - x_j| \leq \varepsilon_n(x_k)$ such that

$$\left| |x_i - x_j| - D_{ik} g_n \right| \leq \varepsilon_n(x).$$

At this stage, a difference arises. The proof of (Hashimoto et al., 2015, Theorem S4.5) bounds the number of steps necessary to reach distance $\varepsilon_n(x_j)$ to the target, but for a general choice of $h(x)$ this does not guarantee that we can reach $x_j$.

For general $h(x)$, we instead show that two extra jumps are sufficient. Because $h(1) > 0$ and $h$ is continuous at 1, there exists some interval $(c_1, 1)$ and some $c_2 > 0$ such that

$$\inf_{x \in (c_1, 1)} h(x) > c_2.$$

This annulus will yield a lower bound on the true connectivity. If $|x_i - x_j| \leq \varepsilon_n(x_i)$, then the probability that there is some point $x_k$ such that the path $x_i \to x_k \to x_j$ exists in $G_n$ is governed by

$$\mathbb{P}(D_{ij} > 2) = (1 - c_2)^{2|\mathsf{NB}_n(x_i) \cap \mathsf{NB}^{in}_n(x_j)|}$$

where

$$|\mathsf{NB}_n(x_i) \cap \mathsf{NB}^{in}_n(x_j)| \sim \text{Pois}(g_n^d n \tau(x_i - x_j))$$

and $\tau(z)$ is the total overlapping density between the connectivity kernel of $x_i$ and $x_j$. This is lower bounded by the annulus; for any $d > 2$ the annuli have nonzero overlap volume and

$$\tau(z) \geq c_2^2 \int_{x \in B(0,1)} 1_{1 > |x| > c_1} 1_{1 > |1 - x| > c_1} dx \geq 0.$$

This implies that $|\mathsf{NB}_n(x_i) \cap \mathsf{NB}_n^{in}(x_j)| = \Theta(k)$ with high probability and therefore

$$\mathbb{P}(D_{ij} > 2) = (1 - c_2)^{2|\mathsf{NB}_n(x_i) \cap \mathsf{NB}_n^{in}(x_j)|} \to 0.$$

Thus, there exists a two step path from $x_i$ to $x_j$ whenever $|x_i - x_j| < \varepsilon_n(x_i)$. Combined with the analogue of (Hashimoto et al., 2015, Theorem S4.5), this shows that with high probability there is a walk of $\widehat{\varepsilon}$-weight at most $|x_i - x_k| + 2\varepsilon_n(x_k)$ from $x_i$ to $x_k$. We conclude in the same way as in the constant kernel case. $\qquad\square$

We now require a lemma on the continuity of functions on Skorokhod space. For this, we recall the metric which induces the relevant topology on Skorokhod space. Let $\Lambda$ be the set of strictly increasing continuous bijections $[0, \infty) \to [0, \infty)$. The Skorokhod metric on $\mathsf{D}([0, \infty), \overline{D})$ and $\mathsf{D}([0, \infty), \mathbb{R}_{\geq 0})$ is given by

$$\sigma(f, g) = \inf_{\lambda \in \Lambda} \max\{||\lambda - \mathrm{id}||, ||f - g \circ \lambda||\},$$

where $|| \cdot ||$ denotes the sup-norm on the relevant space.

**Lemma S6.4.** *Let $B \subset D$ be any ball and $\overline{T}_B^x$ the hitting time from $x$ to $B$ of Brownian motion with reflecting boundary condition in $D$. As a map $\mathsf{D}([0, \infty), \overline{D}) \to \mathbb{R}_{\geq 0}$, the hitting time $\overline{T}_B^x$ is continuous on the subset of $\mathsf{C}([0, \infty), \overline{D})$ of paths whose hitting time to $B$ is finite.*

*Proof.* Denote by $\mathsf{C}_B$ the subset of $\mathsf{C}([0, \infty), \overline{D})$ of paths whose hitting time to $B$ is finite. We first claim that the function

$$d_B : \mathsf{D}([0, \infty), \overline{D}) \to \mathsf{D}([0, \infty), \mathbb{R}_{\geq 0})$$

given by composition with the function $\overline{d}_B : \overline{D} \to B$ giving the distance to $B$ is continuous. For any $\varepsilon > 0$, pick $\delta$ by uniform continuity of $\overline{d}_B$ so that $\delta < \varepsilon$ and if $|x - y| < \delta$, then $|\overline{d}_B(x) - \overline{d}_B(y)| < \varepsilon$. If $\sigma(f, g) < \delta$, we have

$$\sigma(d_B(f), d_B(g)) = \inf_{\lambda \in \Lambda} \max\{||\lambda - \mathrm{id}||, ||\overline{d}_B \circ f - \overline{d}_B \circ g \circ \lambda||\}.$$

Because $\sigma(f, g) < \delta$, we may find $\lambda \in \Lambda$ so that $||f - g \circ \lambda|| < \delta$ and $||\lambda - \mathrm{id}|| < \delta$. By our choice of $\delta$, this implies that

$$\max\{||\lambda - \mathrm{id}||, ||\overline{d}_B \circ f - \overline{d}_B \circ g \circ \lambda||\} < \varepsilon$$

and therefore that $\sigma(d_B(f), d_B(g)) < \varepsilon$, establishing continuity.

Now, the image of $\mathsf{C}_B$ under $d_B$ is the subset $\mathsf{C}_0$ of $\mathsf{C}([0, \infty), \mathbb{R}_{\geq 0})$ of paths whose hitting time to 0 is finite. By (Whitt, 1980, Theorem 7.1), the first passage time to 0 is continuous on $\mathsf{C}_0$. The hitting time $\overline{T}_B^x$ is the composition of the first passage time and $d_B$, hence is continuous on $\mathsf{C}_B$ as claimed. $\qquad\square$

**Lemma S6.5.** *Let $B \subset D$ be any ball containing at least one point of $G_n$. For $x_i \in G_n$, let $T_{B,n}^{x_i}$ be the hitting time from $x_i$ to $B$ of the de-biased random walk on $G$. Then $g_n^2 \widehat{T}_{B,n}^{x_i} \xrightarrow{d} \overline{T}_B^x$.*

*Proof.* First, note that both the de-biased random walk and Brownian motion with reflecting boundary condition started at $x_i$ have a.s. finite hitting time to $B$. By Lemma S6.4, the hitting time to $B$ is a.s. continuous on the subset of $\mathsf{D}([0, \infty), \overline{D})$ containing their trajectories. The desired convergence in distribution then follows from Corollary S4.2, the continuous mapping theorem (see (Whitt, 1980, Section 1)), and noting the time-rescaling used in Corollary S4.2. $\qquad\square$

*Proof of Corollary S6.2.* Recall that $T_{B(x_j, p), n}^{x_i}$ is the hitting time of the simple random walk on $G_n$ to $B(x_j, p)$. By Lemma S6.3, for any $\delta > 0$, we have with high probability that

$$T_{B(x_j, s+\delta), n}^{x_i} \leq \widehat{T}_{B(x_j, s), n}^{x_i} \leq T_{B(x_j, s-\delta), n}^{x_i}.$$

Applying Lemma S6.5 to $B(x_j, s \pm \delta)$, we see that

$$g_n^2 T_{B(x_j, s \pm \delta), n}^{x_i} \xrightarrow{d} \overline{T}_{B(x_j, s \pm \delta)}^{x_i},$$

which shows that

$$\overline{T}_{B(x_j, s-\delta)}^{x_i} \leq \lim_{n \to \infty} g_n^2 \widehat{T}_{B(x_j, s), n}^{x_i} \leq \overline{T}_{B(x_j, s+\delta)}^{x_i}$$

for all $\delta > 0$. Sending $\delta \to 0$ yields the result. $\qquad\square$

### 6.3 Proof of Theorem 4.5

We prove here Theorem 6.6. Recall we chose an estimator $\widehat{\varepsilon}(x) \to \overline{\varepsilon}(x)$.

**Theorem S6.6.** *Let $x_i$ and $x_j$ be points in $G_n$ connected by a geodesic not intersecting $\partial D$. For any $\delta > 0$, there exists a choice of $\widehat{\beta}$ and $s > 0$ so that if $\beta = \widehat{\beta} g_n^2$, we have for large $n$ with high probability that*

$$\left| -\log(\mathbb{E}[\exp(-\beta \widehat{T}_{B(x_j,s),n}^{x_i})]) / \sqrt{2\widehat{\beta}} - |x_i - x_j| \right| < \delta.$$

Our proof will proceed by converting to the continuous setting by Corollary 6.2 and then reducing to the case of Brownian motion without boundary which was analyzed in Lemma 6.1. Because we are in the setting of Brownian motion with reflecting boundary conditions, we must apply the "principle of not feeling the boundary" to show that our results are unaffected by it. For this, we define some events to condition on.

Let $\mathcal{G}$ be the geodesic from $x_i$ to $x_j$, and for a distance scale $\rho$, let $\mathcal{G}(\rho)$ be the set of all points of distance less than $\rho$ from $\mathcal{G}$. Choose $\rho$ small enough so that $\mathcal{G}(\rho) \subset D$. For a distance $s > 0$, let $B_t$ be a Brownian motion without boundary started at $x_i$, and let $\overline{T}_{B(x_j,s)}^{x_i}$ be its hitting time to $B(x_j, s)$. For a time $t^\star > 0$, define the following events:

- let $E_1$ be the event that $\overline{T}_{B(x_j,s)}^{x_i} < t^\star$;

- let $E_2$ be the event that $E_1$ holds and $B_t$ hits $B(x_j, s)$ before $\mathcal{G}(\rho)$;

- let $E_3$ and $E_4$ denote the analogous events for Brownian motion with boundary.

Notice that $\mathbb{P}(E_2) = \mathbb{P}(E_4)$. In the rest of this section, we will consider the scalings $t^\star = s^\gamma$ and $\widehat{\beta} = s^\alpha$ for some $\gamma > 0$ and $\alpha < 0$ so that $\alpha + \gamma > 0$, so that $\widehat{\beta} t^\star \to \infty$ as $s \to 0$.

Let $p_t^R(x,y)$, $p_t^K(x,y)$, $p_t^G(x,y)$, and $p_t^F(x,y)$ be the transition density of Brownian motion started at $x$ and run for time $t$ with reflecting boundary condition, killed at $\partial D$, killed at $\partial G(\rho)$, and no boundary condition, respectively. For $\star \in \{R, K, G, F\}$, let $h^\star(T)$ be the probability that respective process hits $B(x_j, s)$ before time $T$, and let $h^\star(t, x)$ be the density of hitting at $x \in B(x_j, s)$ at time $t$. Note that $p_t^K(x,y) \le p_t^R(x,y)$, $p_t^K(x,y) \le p_t^F(x,y)$, and $p_t^G(x,y) \le p_t^F(x,y)$. We have the following three lemmas, which are instances of "the principle of not feeling the boundary."

**Lemma S6.7.** *For $x, y$ a distance at least $\rho' > 0$ to $\partial G(\rho)$, there are constants $t_0 > 0$ and $\lambda > 0$ dependent only on $\rho$ so that for $t < t_0$, we have*

$$\frac{p_t^G(x,y)}{p_t^F(x,y)} \ge 1 - e^{-\lambda t^{-1}}.$$

*Proof.* This follows from (Hsu, 1995, Theorem 1.2) and the results of (Varadhan, 1967). □

**Lemma S6.8.** *For $x, y$ a distance at least $\rho' > 0$ to $\partial D$, there are constants $t_0 > 0$ and $\lambda > 0$ dependent only on $\rho$ so that for $t < t_0$, we have*

$$\frac{p_t^K(x,y)}{p_t^F(x,y)} \ge 1 - e^{-\lambda t^{-1}}.$$

*Proof.* This follows from (Hsu, 1995, Theorem 1.2) and the results of (Varadhan, 1967). □

**Lemma S6.9.** *For $x, y$ a distance at least $\rho' > 0$ to $\partial D$, there are constants $t_0 > 0$ and $\lambda > 0$ dependent only on $\rho$ so that for $t < t_0$, we have*

$$\frac{p_t^K(x,y)}{p_t^R(x,y)} \ge 1 - e^{-\lambda t^{-1}}.$$

*Proof.* Note that our domain $D$ is a Lipschitz domain in the sense of (Bass & Hsu, 1991, Section 3). Therefore, by (Bass & Hsu, 1991, Theorem 3.1, Theorem 3.4, and Remark 3.11), the reflecting Brownian motion in $D$ has transition density $p_t^R(x,y)$ satisfying

$$C_1 t^{-d/2} e^{-c_1 \frac{|x-y|^2}{t}} \le p_t^R(x,y) \le C_2 t^{-d/2} e^{-c_2 \frac{|x-y|^2}{t}} \tag{11}$$

for constants $c_1, c_2, C_1, C_2$ and small enough $t$. This verifies the conditions of (Hsu, 1995, Theorem 1.2), yielding the conclusion. □

We now prove a general lemma on when the probability of hitting $B(x_j, s)$ before a time $t^\star$ is asymptotically equal for two processes.

**Lemma S6.10.** *Let $Q$ be a diffusion process with transition densities $p_t^Q(x, y)$, and let $p_t^K(x, y)$ of $Q$ killed at some boundary. If for some $\lambda > 0$ and small enough $s$, we have for all $t < t^\star$ and $x, y \in B$ that*

$$1 \geq \frac{p_t^K(x_i, x)}{p_t^Q(x_i, x)} \geq 1 - e^{-\lambda t^{-1}} \ \text{ and } \ 1 \geq \frac{p_t^K(x, y)}{p_t^Q(x, y)} \geq 1 - e^{-\lambda t^{-1}},$$

*then the probabilities $h^K(t^\star)$ and $h^Q(t^\star)$ that $K$ and $Q$ hit $B(x_j, s)$ before $t^\star$ are asymptotically equal.*

*Proof.* Let $B = B(x_j, s)$, and consider $s$ small enough so that $B(x_j, s) \subset D$. For $x \in B$ and $t > 0$, let $h^K(t, x)$ and $h^Q(t, x)$ be the densities of the first passage time to $B$, and let $h^K(T)$ and $h^Q(T)$ be the probabilities that the respective first passage times are at most $T$. Note that $h^K(t, x) \leq h^Q(t, x)$. For $\star \in \{K, Q\}$, we have

$$h^\star(t, x) = p_t^\star(x_i, x) - \int_0^t \int_{y \in B} p_{t-\tau}^\star(y, x) h^\star(\tau, y) \, dy \, d\tau$$

so we may integrate to obtain

$$h^\star(T) = \int_0^T \int_{x \in B} p_t^\star(x_i, x) \, dt \, dx - \int_0^T \int_0^t \int_{x, y \in B} p_{t-\tau}^\star(y, x) h^\star(\tau, y) \, dy \, dx \, d\tau \, dt. \tag{12}$$

Define the differences $d(T) := h^Q(T) - h^K(T)$, $d(t, x) = h^Q(t, x) - h^K(t, x)$, and $e_t(x, y) := p_t^Q(x, y) - p_t^K(x, y)$. By assumption, if $x = x_i$ or $x, y \in B$, we have

$$e_t(x, y) \leq e^{-\lambda t^{-1}} p_t^Q(x, y).$$

Subtracting (12) for $\star \in \{K, Q\}$, we obtain

$$
\begin{aligned}
d(T) &= \int_0^T \int_{x \in B} e_t(x_i, x) \, dt \, dx + \int_0^T \int_0^t \int_{x, y \in B} e_{t-\tau}(y, x) h^K(\tau, y) \, dy \, dx \, d\tau \, dt \\
&\quad + \int_0^T \int_0^t \int_{x, y \in B} p_{t-\tau}^K(y, x) d(\tau, y) \, dy \, dx \, d\tau \, dt \\
&\leq \int_0^T \int_{x \in B} e^{-\lambda t^{-1}} p_t^Q(x_i, x) \, dt \, dx + \int_0^T \int_0^t \int_{x, y \in B} e^{-\lambda(t-\tau)^{-1}} p_{t-\tau}^Q(y, x) h^K(\tau, y) \, dy \, dx \, d\tau \, dt \\
&\quad + \int_0^T \int_0^t \int_{y \in B} d(\tau, y) \, d\tau \, dt \, dy \\
&\leq \int_0^T e^{-\lambda t^{-1}} \, dt + \int_0^T \int_0^t \int_{y \in B} e^{-\lambda(t-\tau)^{-1}} h^K(\tau, y) \, dy \, d\tau \, dt + \int_0^T d(\tau) \, d\tau \\
&\leq 2 \int_0^T e^{-\lambda t^{-1}} \, dt + \int_0^T d(\tau) \, d\tau \\
&\leq 2 T e^{-\lambda T^{-1}} + \int_0^T d(\tau) \, d\tau.
\end{aligned}
$$

By Gronwall's inequality, this implies that

$$d(T) \leq 2 T e^{-\lambda T^{-1}} + 2 \int_0^T \tau e^{-\lambda \tau^{-1}} (T - \tau) \, d\tau \leq 2(T + T^3) e^{-\lambda T^{-1}}.$$

We conclude that

$$\lim_{s \to \infty} d(t^\star) = \lim_{s \to \infty} h^Q(t^\star) - h^K(t^\star) = 0. \qquad \square$$

**Lemma S6.11.** *As $s \to 0$, we have $\mathbb{P}(E_2 \mid E_1) \to 1$.*

*Proof.* Let $B = B(x_j, s)$. By Lemma 6.7 with $\rho'$ small enough, we have for some $\lambda > 0$ and small enough $s$ that for all $t < t^\star$ and $x, y \in B$ that

$$\frac{p_t^G(x_i, x)}{p_t^F(x_i, x)} \geq 1 - e^{-\lambda t^{-1}} \text{ and } \frac{p_t^G(x, y)}{p_t^F(x, y)} \geq 1 - e^{-\lambda t^{-1}}.$$

Notice that $\mathbb{P}(E_2)$ is the probability that the Brownian motion killed at $G(\rho)$ hits $B$ before $t^\star$ and $\mathbb{P}(E_1)$ is the probability that the free Brownian motion hits $B$ before $t^\star$. Therefore, Lemma 6.10 implies that

$$\lim_{s \to 0} \mathbb{P}(E_1) = \lim_{s \to 0} \mathbb{P}(E_2),$$

from which we conclude that

$$\lim_{s \to 0} \mathbb{P}(E_2 \mid E_1) = \lim_{s \to 0} \frac{\mathbb{P}(E_2)}{\mathbb{P}(E_1)} = 1. \qquad \square$$

**Lemma S6.12.** *As $s \to 0$, we have $\mathbb{P}(E_4 \mid E_3) \to 1$.*

*Proof.* Applying Lemma 6.10 twice using Lemmas 6.9 and 6.8 implies that

$$\lim_{s \to \infty} \mathbb{P}(E_1) = \lim_{s \to \infty} h^F(t^\star) = \lim_{s \to \infty} h^K(t^\star) = \lim_{s \to \infty} h^R(t^\star) = \lim_{s \to \infty} \mathbb{P}(E_3).$$

We conclude from Lemma 6.11 that

$$\lim_{s \to \infty} \mathbb{P}(E_4 \mid E_3) = \lim_{s \to \infty} \frac{\mathbb{P}(E_4)}{\mathbb{P}(E_3)} = \lim_{s \to \infty} \frac{\mathbb{P}(E_2)}{\mathbb{P}(E_1)} = \lim_{s \to \infty} \mathbb{P}(E_2 \mid E_1) = 1. \qquad \square$$

*Proof of Theorem 6.6.* Throughout this proof, we will take $t^\star = s^\gamma$ and $\widehat{\beta} = s^\alpha$ for some fixed $\alpha < 0$ and $\gamma > 0$ so that $\alpha + \gamma > 0$. We will pick a small $s > 0$ at the end of the proof.

**Bounding the effect of conditioning on $E_4$ on the process with boundary:** By Corollary 6.2, for any $\widehat{\beta}$ we have that

$$-\log(\mathbb{E}[\exp(-\widehat{\beta} g_n^2 \widehat{T}_{B(x_j, s), n}^{x_i})]) / \sqrt{2\widehat{\beta}} \xrightarrow{d} -\log(\mathbb{E}[\exp(-\widetilde{\beta} \overline{T}_{B(x_j, s)}^{x_i})]) / \sqrt{2\widehat{\beta}}.$$

Conditioning on $E_3$ and $E_4$, we see that

$$\begin{aligned}
\mathbb{E}[\exp(-\widetilde{\beta} \overline{T}_{B(x_j, s)}^{x_i})] = {} & \mathbb{E}[\exp(-\widehat{\beta} \overline{T}_{B(x_j, s)}^{x_i}) \mid E_3^c] \, \mathbb{P}(E_3^c) \\
& + \mathbb{E}[\exp(-\widehat{\beta} \overline{T}_{B(x_j, s)}^{x_i}) \mid E_4] \, \mathbb{P}(E_4) \\
& + \mathbb{E}[\exp(-\widehat{\beta} \overline{T}_{B(x_j, s)}^{x_i}) \mid E_3 \cap E_4^c] \, \mathbb{P}(E_4^c \mid E_3) \, \mathbb{P}(E_3).
\end{aligned}$$

By definition of $E_3$, we have $0 \leq \mathbb{E}[\exp(-\widehat{\beta} \overline{T}_{B(x_j, s)}^{x_i}) \mid E_3^c] \, \mathbb{P}(E_3^c) \leq e^{-\widehat{\beta} t^\star}$. By the trivial bound $\exp(-\widehat{\beta} \overline{T}_{B(x_j, s)}^{x_i}) \leq 1$, we find that

$$0 \leq \mathbb{E}[\exp(-\widehat{\beta} \overline{T}_{B(x_j, s)}^{x_i}) \mid E_3 \cap E_4^c] \, \mathbb{P}(E_4^c \mid E_3) \, \mathbb{P}(E_3) \leq 1 - \mathbb{P}(E_4 \mid E_3).$$

By Lemma 6.12, for any $\tau > 0$, for small enough $s > 0$ we have $e^{-\widehat{\beta} t^\star} < \tau$ and $1 - \mathbb{P}(E_4 \mid E_3) < \tau$. Noting also that $\mathbb{P}(E_4) = \mathbb{P}(E_2)$ and $\mathbb{E}[\exp(-\widehat{\beta} \overline{T}_{B(x_j, s)}^{x_i}) \mid E_4] = \mathbb{E}[\exp(-\widehat{\beta} \widehat{T}_{B(x_j, s)}^{x_i}) \mid E_2]$, we conclude for small enough $s$ that

$$\left| \mathbb{E}[\exp(-\widehat{\beta} \overline{T}_{B(x_j, s)}^{x_i})] - \mathbb{E}[\exp(-\widehat{\beta} \widehat{T}_{B(x_j, s)}^{x_i}) \mid E_2] \, \mathbb{P}(E_2) \right| < 2\tau. \tag{13}$$

**Bounding the effect of conditioning on $E_2$ on the process without boundary:** We now compare to the computations for Brownian motion without boundary. By conditioning on $E_1$ and $E_2$, we have that

$$\begin{aligned}
\mathbb{E}[\exp(-\widehat{\beta} \widehat{T}_{B(x_j, s)}^{x_i})] = {} & \mathbb{E}[\exp(-\widehat{\beta} \widehat{T}_{B(x_j, s)}^{x_i}) \mid E_1^c] \, \mathbb{P}(E_1^c) \\
& + \mathbb{E}[\exp(-\widehat{\beta} \widehat{T}_{B(x_j, s)}^{x_i}) \mid E_2] \, \mathbb{P}(E_2) \\
& + \mathbb{E}[\exp(-\widehat{\beta} \widehat{T}_{B(x_j, s)}^{x_i}) \mid E_1 \cap E_2^c] \, (1 - \mathbb{P}(E_2 \mid E_1)) \, \mathbb{P}(E_1).
\end{aligned}$$

We again note that

$$0 \leq \mathbb{E}[\exp(-\widehat{\beta}\widehat{T}^{x_i}_{B(x_j,s)}) \mid E_1^c]\,\mathbb{P}(E_1^c) \leq e^{-\widehat{\beta}t^\star}$$

and

$$0 \leq \mathbb{E}[\exp(-\widehat{\beta}\widehat{T}^{x_i}_{B(x_j,s)}) \mid E_1 \cap E_2^c]\,(1 - \mathbb{P}(E_2 \mid E_1))\,\mathbb{P}(E_1) \leq 1 - \mathbb{P}(E_2 \mid E_1).$$

These together with Lemma 6.11 imply that for any $\tau > 0$, for small enough $s > 0$ we have that $e^{-\widehat{\beta}t^\star} < \tau$ and $1 - \mathbb{P}(E_2 \mid E_1) < \tau$. We conclude for small enough $s$ that

$$\left| \mathbb{E}[\exp(-\widehat{\beta}\widehat{T}^{x_i}_{B(x_j,s)})] - \mathbb{E}[\exp(-\widehat{\beta}\widehat{T}^{x_i}_{B(x_j,s)}) \mid E_2]\,\mathbb{P}(E_2) \right| < 2\tau. \tag{14}$$

Combining (13) and (14), we conclude for small enough $s$ that

$$\left| \mathbb{E}[\exp(-\widehat{\beta}\overline{T}^{x_i}_{B(x_j,s)})] - \mathbb{E}[\exp(-\widehat{\beta}\widehat{T}^{x_i}_{B(x_j,s)})] \right| < 4\tau. \tag{15}$$

**Aggregating the estimates:** To conclude, for any $\delta > 0$ and $\widehat{\beta}_0 > 0$, choose $\tau > 0$ small enough so that if $|x - y| < 4\tau$, then for all $\widehat{\beta} > \widehat{\beta}_0$, we have

$$\left| \log(x)/\sqrt{2\widehat{\beta}} - \log(y)/\sqrt{2\widehat{\beta}} \right| < \delta/3.$$

Now, choose $s > 0$ small enough and $n$ large enough so that $\widehat{\beta} > \widehat{\beta}_0$, and for this $\tau$, we have:

- by our previous discussion, (15) holds;

- by Lemma S6.1, we have

$$\left| -\log(\mathbb{E}[\exp(-\widehat{\beta}\widehat{T}^{x_i}_{B(x_j,s)})])/\sqrt{2\widehat{\beta}} - |x_i - x_j| \right| < \delta/3;$$

- by Corollary S6.2, we have

$$\left| -\log(\mathbb{E}[\exp(-\widehat{\beta}\overline{T}^{x_i}_{B(x_j,s)})])/\sqrt{2\widehat{\beta}} + \log(\mathbb{E}[\exp(-\widehat{\beta}g_n^2\widehat{T}^{x_i}_{B(x_j,s),n})])/\sqrt{2\widehat{\beta}} \right| < \delta/3.$$

For these choices of $\tau$, $s$, and $n$, we have by (15) that

$$\left| \log(\mathbb{E}[\exp(-\widehat{\beta}\overline{T}^{x_i}_{B(x_j,s)})])/\sqrt{2\widehat{\beta}} - \log(\mathbb{E}[\exp(-\widehat{\beta}\widehat{T}^{x_i}_{B(x_j,s)})])/\sqrt{2\widehat{\beta}} \right| < \delta/3.$$

Combining the last three inequalities yields the desired

$$\left| \log(\mathbb{E}[\exp(-\widehat{\beta}g_n^2\overline{T}^{x_i}_{B(x_j,s),n})])/\sqrt{2\widehat{\beta}} - |x_i - x_j| \right| < \delta. \qquad \square$$

# 7  1-D bias calculation

We repeat the full theorem statement and proof for the bias characterization.

**Theorem S7.1.** *Let $T^{x_i}_{x_j}$ be the hitting time to $x_j$ of a 1-dimensional Itô process with drift $\mu(x) = \frac{\partial \log(p(x))}{\partial x}\overline{\varepsilon}^2(x)$ and diffusion $\overline{\varepsilon}^2(x)$ started at $x_i$ with reflecting boundary $\gamma$ for $\gamma < x_i < x_j$. The Laplace transform of $T^{x_i}_{x_j}$ admits the asymptotic expansion*

$$\mathbb{E}[-\exp(\beta T^{x_i}_{x_j})] = \frac{c_1}{f(x_i)^{1/4}p(x_i)} \exp\left( -\sqrt{\beta} \int_{x_i}^{x_j} \sqrt{f(s)}ds \right)$$
$$\left( 1 + \left( 1 + o\left( \frac{1}{\sqrt{\beta}} \right) \right) \exp\left( -2\sqrt{\beta} \int_{\gamma}^{x_i} \sqrt{f(x)}dx \right) + o(\exp(-\beta)) \right),$$

*where $f(x) = \frac{2}{\overline{\varepsilon}(x)^2} + \frac{1}{\beta}\frac{\partial \log(p(x))}{\partial x^2} + \frac{1}{\beta}\left( \frac{\partial \log(p(x))}{\partial x} \right)^2$, and $c_1$ is a normalization constant depending on $p$, $\overline{\varepsilon}$, and $j$ to make $E[-\beta T^{x_i}_{x_j}] = 1$.*

*Proof.* Let $\mathbb{E}[\exp(-\beta T_{x_j}^{x_i})] = u(x_i)$, where $u(x)$ is the hitting time to $x_j$ from point $x$. By Feynman-Kac, this is

$$\frac{\partial^2 u}{\partial x^2} + 2\frac{\partial \log(p(x))}{\partial x}\frac{\partial u}{\partial x} + q(x)u = 0,$$

where $q(x) = -2\beta\bar{\varepsilon}(x)^{-2}$. Rewrite this as a perturbation of a second order ODE via the change of variables to obtain

$$y(x) = u(x)\exp\left(\int_\gamma^x \frac{\partial \log(p(y))}{\partial y}dy\right) = u(x)p(x)p(\gamma)^{-1}$$

$$f(x) = \frac{2}{\bar{\varepsilon}(x)^2} + \frac{1}{\beta}\left(\frac{\partial \log(p(x))}{\partial x^2} + \left(\frac{\partial \log(p(x))}{\partial x}\right)^2\right)$$

$$\frac{1}{\beta}\frac{\partial^2 y}{\partial x} = f(x)y(x).$$

Since this is a type of Schrödinger's equation with $f(x) \neq 0$ everywhere we can apply the WKBJ asymptotic expansion (Bender & Orszag, 1999, section 10.1) to obtain

$$y(x) = \frac{c_1}{f(x)^{1/4}}\exp\left(-\sqrt{\beta}\int_{x_0}^x \sqrt{f(s)}ds\right) + \frac{c_2}{f(x)^{1/4}}\exp\left(\sqrt{\beta}\int_{x_0}^x \sqrt{f(s)}ds\right) + o(\exp(-\beta)).$$

Since we assumed $x_i < x_j$ and by the boundary condition $u(x_j) = 1$ we have

$$u(x) = \frac{c_2 p(\gamma)}{f(x)^{1/4}p(x)}\exp\left(-\sqrt{\beta}\int_x^{x_j} \sqrt{f(s)}ds\right) + \frac{c_1 p(\gamma)}{f(x)^{1/4}p(x)}\exp\left(\sqrt{\beta}\int_x^{x_j} \sqrt{f(s)}ds\right) + o(\exp(-\beta)).$$

To obtain the boundary conditions, note that $u'(\gamma) = 0$. Taking the derivative for $y(x)p(x)$, setting to zero and solving for $c_2$ results in

$$c_2 = c_1 \frac{\exp(-2\sqrt{\beta}\int_\gamma^{x_j} \sqrt{f(s)}ds)(p(\gamma)4\sqrt{\beta}f(\gamma)^{3/2} + f'(\gamma)) - f(\gamma)p'(\gamma)}{4\sqrt{\beta}f(\gamma)^{3/2}p(\gamma) - p(\gamma)f'(\gamma) + 4f(\gamma)p'(\gamma)} + o(\exp(-\beta)),$$

from which we obtain

$$c_2 = c_1 \exp\left(-2\sqrt{\beta}\int_\gamma^{x_j} \sqrt{f(s)}ds\left(1 + o\left(\sqrt{\frac{1}{\beta}}\right)\right)\right).$$

Pulling out the $-\sqrt{\beta}$ term, we get

$$u(x_i) = \mathbb{E}[\exp(-\beta T_{x_j}^{x_i})] = \frac{c_1 p(\gamma)}{f(x_i)^{1/4}p(x_i)}\exp\left(-\sqrt{\beta}\int_{x_i}^{x_j} \sqrt{f(s)}ds\right)$$

$$\left(1 + \left(1 + o\left(\frac{1}{\sqrt{\beta}}\right)\right)\exp\left(-2\sqrt{\beta}\int_\gamma^{x_i} \sqrt{f(x)}dx\right) + o(\exp(-\beta))\right). \quad \square$$

We now connect this statement to the discrete walk.

**Corollary S7.2.** *Let $T_{B(x_j,s),n}^{x_i}$ be the discrete hitting time to a $s$ ball around $x_j$ where $s$ is selected as given in Theorem S2.13. Then the simple random walk over a graph constructed on density $p(x)$ and scale $\bar{\varepsilon}(x)$ has the following log-LTHT under the boundary conditions of Theorem S4.6*

$$-\log(\mathbb{E}[\exp(-\beta T_{B(x_j,s),n}^{x_i} g_n^2)])/\sqrt{2\beta} \to \int_{x_i}^{x_j} \sqrt{\frac{1}{\bar{\varepsilon}(x)} + \frac{1}{\beta}\left(\frac{\partial \log(p(x))}{\partial x^2} + \left(\frac{\partial \log(p(x))}{\partial x}\right)^2\right)}dx$$

$$+ \frac{\log(p(x_i)/p(x_j)) + \log(f(x_i)/f(x_j))/4}{\sqrt{2\beta}} + o(\log(1 + e^{-\sqrt{2\beta}})/\sqrt{2\beta}).$$

*Proof.* Taking the logarithm of the result of Theorem 7.1 and noting the initial condition $u(x_j) = 1$ implies that asymptotically we have

$$c_1 \propto \left( \frac{1}{f(x_j)^{1/4} p(x_j)} (1 + o(e^{-2\sqrt{\beta}})) \right)^{-1} \to f(x_j)^{1/4} p(x_j),$$

which completes the continuous statement. The convergence of the hitting time to its discrete counterpart follows from Theorem S2.13. □

## 8 Basic noise resistance

We give details for the basic noise bound from the main text footnote. Our goal is to prove the following statement about random walks.

**Theorem S8.1.** *Let $G_n$ be generated by the noise model of definition 4.7 with $\sum_j q_j = o(g_n^2)$. Then the simple random walk over $G_n$ converges to the same limit as the noiseless case in Theorem 2.2.*

*Proof.* Since the boundaries of both noisy and noiseless graphs are identical, we need only verify the moment conditions in the proof of Theorem 2.2. In particular we require that under any noise $q$, we have

$$\lim_{n \to \infty} g_n^{-2} \mathbb{E}[X_{t+1}^n - X_t^n | X_t^n] = \nabla \log(p(X_t^n)) \bar{\varepsilon}(X_t^n)^2$$

$$\lim_{n \to \infty} g_n^{-2} \mathsf{Cov}[X_{t+1}^n | X_t^n] = \bar{\varepsilon}(X_t^n)^2 \cdot I_n$$

$$\lim_{n \to \infty} g_n^{-2} \mathbb{E}[| X_{t+1}^n - X_t^n |^{2+\alpha} | X_t^n] = 0,$$

which we show in the Lemma S8.2 and S8.3 below. By the Stroock-Varadhan criterion, this implies convergence to Theorem 2.2, as well as any macroscopic quantities such as hitting times, or LTHTs with $\beta = \Theta(g_n^2)$. □

We now prove the moment bounds required for convergence of the noisy graph.

**Lemma S8.2** (Noisy moments). *If the noisy graph $G_n$ is generated by the noise model of Definition 4.7, for any choice of latent noise parameters $q_j$ such that $\sum_j q_j = o(g_n^2)$ then we have for $\alpha > 0$ that*

$$\lim_{n \to \infty} g_n^{-2} \mathbb{E}[X_{t+1}^n - X_t^n | X_t^n] = \nabla \log(p(X_t^n)) \bar{\varepsilon}(X_t^n)^2$$

$$\lim_{n \to \infty} g_n^{-2} \mathsf{Cov}[X_{t+1}^n | X_t^n] = \bar{\varepsilon}(X_t^n)^2 \cdot I_n$$

$$\lim_{n \to \infty} g_n^{-2} \mathbb{E}[| X_{t+1}^n - X_t^n |^{2+\alpha} | X_t^n] = 0.$$

*Proof.* Let $\overline{X}$ denote quantities in the noise-free graph. We recall from (Hashimoto et al., 2015, Theorem 3.3) that

$$\lim_{n \to \infty} g_n^{-2} \mathbb{E}[\overline{X}_{t+1}^n - \overline{X}_t^n | \overline{X}_t^n = x] = \nabla \log(p(x)) \bar{\varepsilon}(x)^2$$

$$\lim_{n \to \infty} g_n^{-2} \mathsf{Cov}[\overline{X}_{t+1}^n | \overline{X}_t^n = x] = \bar{\varepsilon}(x)^2 \cdot I_n$$

$$\lim_{n \to \infty} g_n^{-2} \mathbb{E}[| \overline{X}_{t+1}^n - \overline{X}_t^n |^{2+\alpha} | \overline{X}_t^n = x] = 0.$$

Let $\hat{q} = \sum_i q_i$ so that $\hat{q} = o(g_n^2)$. In the noisy graph, we first check the expectation via

$$\lim_{n \to \infty} g_n^{-2} \mathbb{E}[X_{t+1}^n - X_t^n \mid X_t^n = x] = \lim_{n \to \infty} (1 - \hat{q}) g_n^{-2} \mathbb{E}[\overline{X}_{t+1}^n - \overline{X}_t^n \mid \overline{X}_t^n = x] + g_n^{-2} \sum_i q_i (x_i - x)$$

$$= \lim_{n \to \infty} g_n^{-2} \mathbb{E}[\overline{X}_{t+1}^n - \overline{X}_t^n | \overline{X}_t^n = x]$$

$$= \nabla \log(p(x)) \bar{\varepsilon}(x)^2.$$

The covariance follows because for all indices $i$ and $j$ we have

$$\lim_{n\to\infty} g_n^{-2}\mathbb{E}[(X_{t+1}^n - X_t^n)_i(X_{t+1}^n - X_t^n)_j \mid X_t^n = x]$$
$$= \lim_{n\to\infty}(1-\widehat{q})g_n^{-2}\mathbb{E}[(X_{t+1}^n - X_t^n)_i(X_{t+1}^n - X_t^n)_j \mid \overline{X}_t^n = x] + g_n^{-2}\sum_k q_k(x_k - x)_i(x_k - x)_j$$
$$= \lim_{n\to\infty}g_n^{-2}\mathbb{E}[(\overline{X}_{t+1}^n - \overline{X}_t^n)_i(\overline{X}_{t+1}^n - \overline{X}_t^n)_j \mid \overline{X}_t^n = x]$$
$$= \delta_{ij}\overline{\varepsilon}(x)^2.$$

Finally, the higher moments follow because we have

$$\lim_{n\to\infty} g_n^{-2}\mathbb{E}[|X_{t+1}^n - X_t^n|^{2+\alpha} \mid X_t^n = x]$$
$$= \lim_{n\to\infty}g_n^{-2}(1-\widehat{q})g_n^{-2}\mathbb{E}[|\overline{X}_{t+1}^n - \overline{X}_t^n|^{2+\alpha} \mid \overline{X}_t^n = x] + g_n^{-2}\sum_i q_i|x_i - x|^{2+\alpha}$$
$$= \lim_{n\to\infty}g_n^{-2}(1-\widehat{q})g_n^{-2}\mathbb{E}[|\overline{X}_{t+1}^n - \overline{X}_t^n|^{2+\alpha} \mid \overline{X}_t^n = x]$$
$$= 0,$$

where we use that $|x_i - x|^{2+\alpha} = O(1)$. $\qquad\square$

**Lemma S8.3** (Strong LLN for noisy moments). *For a function $f(x)$ such that $\sup_{x\in B(0,\varepsilon)}|f(x)| < \varepsilon$ and $\sup_{x\in D}|f(x)| < C$ for some constant $C$, given $(\star)$ we have uniformly in $x\in\mathcal{X}_n$ that*

$$g_n^{-2}\sum_{y\in\mathsf{NB}_n(x)}\frac{1}{|\mathsf{NB}_n(x)|}f(y-x) \overset{a.s.}{\to} g_n^{-2}\int_{y\in B(x,\varepsilon_n(x))}f(y-x)\frac{p(y)}{p_{\varepsilon_n(x)}(x)}dy.$$

*Proof.* Denote the claimed value of the limit by $\mu(x)$. Let the set of non-noise out-neighbors of $x$ be $\overline{\mathsf{NB}}_n(x)$ and the set of noise out-neighbors of $x$ be $\widetilde{\mathsf{NB}}_n(x)$, where we consider noise edges to be strictly non-geometric edges. We have uniformly in $x\in cX$ that

$$g_n^{-2}\frac{\sum_{y\in\overline{\mathsf{NB}}_n(x)}f(y-x) + \sum_{y\in\widetilde{\mathsf{NB}}_n(x)}f(y-x)}{|\overline{\mathsf{NB}}_n(x)| + |\widetilde{\mathsf{NB}}_n(x)|} = g_n^{-2}\frac{\sum_{y\in\overline{\mathsf{NB}}_n(x)}f(y-x) + o(Cg_n^2)}{|\overline{\mathsf{NB}}_n(x)| + o(g_n^2)}$$
$$\overset{a.s.}{\to} g_n^{-2}\sum_{y\in\overline{\mathsf{NB}}_n(x)}\frac{f(y-x)}{|\overline{\mathsf{NB}}_n(x)|},$$

so the result follows by the noise-less result in (Hashimoto et al., 2015). $\qquad\square$

Now the behavior of noisy hitting times can be recovered by combining Lemma S8.1 with the convergence result of Corollary S6.2.

**Theorem S8.4.** *Let $G_n$ be a noisy geometric graph with noise $\sum_j q_j = o(g_n^2)$. For any $\delta$, there exists some $\beta = \widehat{\beta}g_n^2$, $s$, $c$ such that*

$$\left| -\frac{\log(\mathbb{E}[\exp(-\beta T_{B(x_j,s),n}^{x_i})])}{\sqrt{2\beta}}g_n - c|x_i - x_j| \right| \leq \delta$$

*with high probability as $n\to\infty$.*

*Proof.* By Lemma S8.1, the noisy and noise-free walks converge to the same continuum limit, and this guarantees that by Corollary S6.2 that their hitting times converge in distribution. Applying Theorem 4.5 gives the desired result. $\qquad\square$

This is a basic, but useful result for robustness of hitting times. Up to $o(1)$ noise edges can be allowed for each vertex without disrupting the global convergence of hitting times.

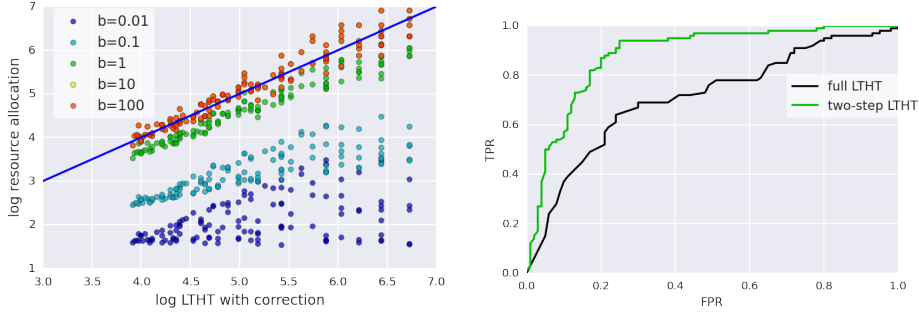

(a) Modified LTHT rapidly converges to RA index.

(b) Conditioning on $t > 1$ substantially outperforms naive LTHT.

# 9 Resource allocation index

Recall that the directed RA index was defined by

$$R_{ij} = \sum_{x_k \in \mathsf{NB}_n(x_i) \cap \mathsf{NB}_n^{\text{in}}(x_j)} \frac{1}{|\mathsf{NB}_n(x_k)|}$$

and the modified log-LTHT was defined by

$$M_{ij}^{\text{mod}} = -\log(\mathbb{E}[\exp(-\beta T_{x_j,n}^{x_i}) \mid T_{x_j,n}^{x_i} > 1]).$$

## 9.1 RA index reduction

**Theorem S9.1.** *If $\beta = \omega(\log(g_n^d n))$ and $x_i$ and $x_j$ have at least one common neighbor, then*

$$M_{ij}^{mod} - 2\beta \to -\log(R_{ij}) + \log(|\mathsf{NB}_n(x_i)|).$$

*Proof.* Let $P_{ij}(t)$ be the probability of going from $x_i$ to $x_j$ in $t$ steps, and $H_{ij}(t)$ the probability of not hitting before time $t$. Factoring the two-step hitting time yields

$$M_{ij}^{\text{mod}} = 2\beta - \log(P_{ij}(2)) - \log\Big(1 + \sum_{t=3}^{\infty} \frac{P_{ij}(t)}{P_{ij}(2)} H_{ij}(t) e^{-\beta(t-2)}\Big).$$

Let $k_{\max}$ be the maximal out-degree which occurs in $G_n$. By assumption, at least one of the at most $k_{\max}^2$ two-step paths from $x_i$ goes to $x_j$, we have the bound $\frac{P_{ij}(t)}{P_{ij}(2)} \leq k_{\max}^2$. For $\beta = \omega(\log(g_n^d n))$, we see that $\beta = \omega(2\log(k_{\max}))$ with high probability. Applying the bounds $H_{ij}(t) \leq 1$ and $\frac{P_{ij}(t)}{P_{ij}(2)} \leq k_{\max}^2$, we obtain

$$\sum_{t=3}^{\infty} \frac{P_{ij}(t)}{P_{ij}(2)} H_{ij}(t) e^{-\beta(t-2)} \leq \frac{k_{\max}^2}{e^\beta - 1} = o(k_{\max}^{-1}).$$

We conclude that $M_{ij}^{\text{mod}} \to 2\beta - \log(P_{ij}(2))$. It remains to verify that $\log(P_{ij}(2))$ is related to the resource allocation index by

$$\log(P_{ij}(2)) = \log\left(\frac{1}{|\mathsf{NB}_n(x_i)|} \sum_{k \in \mathsf{NB}_n(x_i) \cap \mathsf{NB}_n^{\text{in}}(x_j)} \frac{1}{\mathsf{NB}_n(x_k)}\right) = \log(R_{ij}) - \log(|\mathsf{NB}_n(x_i)|). \qquad \square$$

## 9.2 RA index robustness

We verify the robustness of the RA index by directly bounding the statistics involved.

**Theorem S9.2.** *If $q_i = q = o(g_n^{d/2})$ for all $i$, then for any $\delta > 0$ there exist cutoffs $c_1, c_2$ and scaling $h_n$ so that with probability at least $1 - \delta$, for any $i, j$ we have*

- $|x_i - x_j| < \min\{\varepsilon_n(x_i), \varepsilon_n(x_j)\}$ *if* $R_{ij} h_n < c_1$;
- $|x_i - x_j| > 2\max\{\varepsilon_n(x_i), \varepsilon_n(x_j)\}$ *if* $R_{ij} h_n > c_2$.

*Proof.* Decompose the out-degree of $x_i$ into expectation and noise terms by

$$|\mathsf{NB}_n(x_i)| = nq + k_i + z_i,$$

where $k_i = \varepsilon_n(x_i)^d p(x_i) V_d n$, $V_d$ is the volume of the $d$-unit ball, and $z_i$ is a random variable giving the remaining error. The number of noise edges has a binomial distribution with $n$ draws and success probability $q$, and the number of geometric edges has a Poisson distribution with rate $k_i$. Therefore, the Chebyshev inequality implies

$$\mathbb{P}(|z_i| > c) \leq \frac{k_i + nq(1-q)}{c^2} < \frac{k_i + nq}{c^2}. \tag{16}$$

Let $\delta_1 = \delta/4$ and define $c$ by the equality

$$\delta_1 = \frac{k_i + nq(1-q)}{c^2} \tag{17}$$

so that $c = \delta_1^{-1/2}\sqrt{k_i + nq}$. For the rest of the proof, we condition on the event that $|z_i| < c$. By Taylor expanding $\frac{1}{|\mathsf{NB}_n(x_i)|}$ in $z_i$, we have that

$$\begin{aligned}
\frac{1}{|\mathsf{NB}_n(x_i)|} &= \frac{1}{nq + k_i} - \frac{z_i}{(nq + k_i)^2} + O\left(\frac{z_i^2}{(nq + k_i)^3}\right) \\
&= \frac{1}{nq + k_i} - O\left(\frac{c}{(nq + k_i)^2}\right).
\end{aligned} \tag{18}$$

By (17), we see that $|z_i| < c$ with probability at least $1 - \delta_1$, which implies that

$$\frac{c}{(nq + k_i)^2} < \delta_1^{-1/2}(nq + k_i)^{-3/2}.$$

By the definition of the RA index, we obtain

$$R_{ij} = \sum_{x_k \in \mathsf{NB}_n(x_i) \cap \mathsf{NB}_n^{in}(x_j)} \left(\frac{1}{nq + k_i} + O(\delta_1^{-1/2}(nq + k_i)^{-3/2})\right).$$

Since our domain is compact, we may define

$$k_n^+ = \sup_x \varepsilon_n(x)^d p(x) V_d n \qquad \text{and} \qquad k_n^- = \inf_x \varepsilon_n(x)^d p(x) V_d n.$$

By construction, $k_n^+ > k_i > k_n^-$ for all $i$. Let $C_{ij} := |\mathsf{NB}_n(x_i) \cap \mathsf{NB}_n^{in}(x_j)|$. Then we have

$$\frac{C_{ij}}{nq + k_n^+} - O\left(\frac{C_{ij}}{\delta_1^{1/2}(nq + k_n^+)^{3/2}}\right) \leq R_{ij} \leq \frac{C_{ij}}{nq + k_n^-} + O\left(\frac{C_{ij}}{\delta_1^{1/2}(nq + k_n^-)^{3/2}}\right). \tag{19}$$

Choose the scaling

$$h_n = \frac{nq + k_n^+}{k_n^+}.$$

We will now bound $C_{ij}$ to control $h_n R_{ij}$. To do this, decompose $C_{ij}$ as

$$C_{ij} = C_{ij}^g + C_{ij}^{n1} + C_{ij}^{n2},$$

where $C_{ij}^g$, $C_{ij}^{n1}$, and $C_{ij}^{n2}$ are defined as follows.

1. Geometric edges ($C_{ij}^g$): If $|x_i - x_j| < \min\{\varepsilon_n(x_i), \varepsilon_n(x_j)\}$ then they share common neighbors due to the geometric graph. Specifically their number of common neighbors has Poisson distribution with mean at least $\tau_d k_i (1 - q)$, where $\tau_d$ is a constant independent of $n$ defined as the overlapping density of two kernels at a unit distance.

2. One noise edge ($C_{ij}^{n1}$): The edge $x_i \to x_k$ occurs by noise but $x_k \to x_j$ is geometric. There are at most $k_n^+$ such vertices with in-edges to $x_j$ and so this is at most a binomial random variable with $k_n^+$ draws and success probability $q$.

3. Two noise edges ($C_{ij}^{n2}$): Both $x_i \to x_k$ and $x_k \to x_j$ may occur by noise, this is at most a binomial random variable with $n - k_n^-$ draws and success probability $q^2$.

**The case of $|x_i - x_j| < \min\{\varepsilon_n(x_i), \varepsilon_n(x_j)\}$:** All types of edges may occur, so we obtain the moment bounds

$$\mathbb{E}\left[\frac{C_{ij}}{k_n^+}\right] \geq \tau_d(1-q)\frac{k_n^-}{k_n^+}$$

$$\mathsf{Var}\left[\frac{C_{ij}}{k_n^+}\right] \leq \frac{\tau_d(1-q)k_n^-}{(k_n^+)^2} + \frac{(n-k_n^-)q^2(1-q^2) + k_n^+ q(1-q)}{(k_n^+)^2} < \frac{\tau_d(1-q)k_n^-}{(k_n^+)^2} + \frac{nq^2 + k_n^+ q}{(k_n^+)^2}.$$

Notice that $\frac{k_n^+}{k_n^-}$ is bounded between the minimum and maximum of $\frac{\varepsilon_n(x)p(x)}{\varepsilon_n(y)p(y)}$ for $x, y \in D$, so $\lim_{n \to \infty} \mathbb{E}\left[\frac{C_{ij}}{k_n^+}\right] \geq c_{ij}$ for some $c_{ij} > 0$. Further, we find that $\mathsf{Var}\left[\frac{C_{ij}}{k_n^+}\right] \to 0$. These imply that for large enough $n$, we have $\frac{C_{ij}}{k_n^+} > c_{ij}$ with probability at least $1 - \delta_1$. Therefore, if $|x_i - x_j| < \min\{\varepsilon_n(x_i), \varepsilon_n(x_j)\}$, we have

$$\lim_{n \to \infty} h_n R_{ij} \geq \lim_{n \to \infty} \frac{C_{ij}}{k_n^+} - O\left(\frac{C_{ij}}{\delta^{1/2}(nq + k_n^+)^{1/2}k_n^+}\right) \geq c_{ij}. \tag{20}$$

**The case of $|x_i - x_j| > 2\max\{\varepsilon_n(x_i), \varepsilon_n(x_j)\}$:** Only noise cases occur, hence we have the moment bounds

$$\mathbb{E}\left[\frac{C_{ij}}{k_n^-}\right] \leq \frac{q^2(n - k_n^-) + qk_n^+}{k_n^-} < \frac{q^2 n}{k_n^-} + \frac{qk_n^+}{k_n^-}$$

$$\mathsf{Var}\left[\frac{C_{ij}}{k_n^-}\right] \leq \frac{(n - k_n^-)q^2(1 - q^2) + k_n^+ q(1 - q)}{(k_n^-)^2} < \frac{qn}{(k_n^-)^2} + \frac{qk_n^+}{(k_n^-)^2}.$$

Because $q = o(g_n^{d/2})$, both the expectation and variance converge to zero and for large enough $n$, we have $C_{ij}/k_n^- \to 0$ probability at least $1 - \delta_1$. Therefore, if $|x_i - x_j| > 2\max\{\varepsilon_n(x_i), \varepsilon_n(x_j)\}$, for we have

$$\lim_{n \to \infty} h_n R_{ij} \leq \lim_{n \to \infty} \frac{C_{ij}(nq + k_n^+)}{k_n^+(nq + k_n^-)} + O\left(\frac{C_{ij}(nq + k_n^+)}{\delta^{1/2}k_n^+(nq + k_n^-)^{3/2}}\right) \to 0. \tag{21}$$

**Combining the cases:** Taking $h_n = \frac{nq + k_n^+}{k_n^+}$, we combine (20) and (21) to conclude that the desired holds with probability at least $1 - 3\delta_1 > 1 - \delta$ for any $c_1 \leq c_{ij}$ and $c_2 > 0$. $\qquad \square$